\documentclass{article}

\usepackage[final]{corl_2025} 

\usepackage{moreverb,url}
\usepackage{xspace}
\usepackage{xcolor}
\usepackage{siunitx}
\usepackage{wrapfig}

\usepackage{amssymb}
\usepackage{amsmath}
\usepackage{amsthm}
\usepackage{amsbsy}
\usepackage{bbm}
\usepackage{dsfont}

\usepackage{graphicx}
\usepackage{wrapfig}
\usepackage{booktabs} 

\usepackage{subfiles}
\usepackage{csquotes}
\usepackage[table]{xcolor} 

\usepackage{annotate-equations}

\usepackage{tocloft}
\usepackage{array}

\usepackage{graphicx}
\usepackage{subfig} 
\usepackage{multirow}
\newcommand\BibTeX{{\rmfamily B\kern-.05em \textsc{i\kern-.025em b}\kern-.08em
T\kern-.1667em\lower.7ex\hbox{E}\kern-.125emX}}

\definecolor{wine}{RGB}{204, 0, 102}
\definecolor{magenta_wine}{RGB}{158, 44, 143}
\definecolor{dusty_wine}{RGB}{143, 59, 101}
\definecolor{ocean}{RGB}{13, 121, 202}
\definecolor{light_ocean}{RGB}{18, 178, 235}
\definecolor{dark_ocean}{RGB}{10, 89, 148}
\definecolor{grey}{RGB}{170, 170, 170}
\definecolor{light-grey}{RGB}{220, 220, 220}
\definecolor{dark_gray}{rgb}{0.2, 0.2, 0.2} 
\definecolor{med-grey}{rgb}{0.3, 0.3, 0.3} 
\definecolor{grape}{RGB}{112,48,160}
\definecolor{aqua}{RGB}{52,172,139}
\definecolor{dark_aqua}{RGB}{35,115,93}
\definecolor{dark_orange}{RGB}{216,92,0}
\definecolor{vibrant_orange}{RGB}{250, 160, 26}
\definecolor{vibrant_blue}{RGB}{14, 120, 255}
\definecolor{vibrant_pink}{RGB}{255, 0, 104}
\definecolor{dark_red}{RGB}{122, 0, 0}
\definecolor{dark_green}{RGB}{0, 92, 34}
\definecolor{dusty_blue}{RGB}{77, 91, 128}
\definecolor{dark_brown}{RGB}{125, 54, 36}
\definecolor{violet}{RGB}{116, 12, 173}

\newcommand{\para}[1]{\smallskip\noindent\textbf{#1. }} 
 
\newcounter{qnum}
\usepackage{pifont}
\newcommand{\cmark}{\ding{51}} 
\newcommand{\xmark}{\ding{55}}%

\newcommand{\state}{s}
\newcommand{\stateSpace}{\mathcal{S}}
\newcommand{\dyns}{f}

\newcommand{\latent}{z}
\newcommand{\latentSpace}{\mathcal{Z}}
\newcommand{\dynz}{f_{\latent}}
\newcommand{\dynamics}{f}
\newcommand{\encoder}{\mathcal{E}}

\newcommand{\ensembleparam}{\psi}

\newcommand{\obs}{o}
\newcommand{\obsSpace}{\mathcal{O}}

\newcommand{\action}{a}
\newcommand{\actionSpace}{\mathcal{A}}

\newcommand{\policy}{\pi}

\newcommand{\policyTask}{\pi^{\text{task}}}

\usepackage{fontawesome5}
\newcommand{\shield}{\text{\tiny{\faShield*}}}

\newcommand{\failure}{\mathcal{F}}
\newcommand{\failureTotal}{\tilde{\failure}}

\newcommand{\marginfunc}{\ell}
\newcommand{\ellz}{\ell_\latent}

\newcommand{\latentfailuremargin}{l}

\newcommand{\valfunc}{V}

\newcommand{\unsafeSet}{\mathcal{U}} 

\newcommand{\monitor}{{\valfunc^{\shield}}} 
\newcommand{\fallback}{\policy^{\shield}} 

\newcommand{\dataset}{\mathcal{D}}
\newcommand{\indist}{\text{WM}}
\newcommand{\ood}{\text{OOD}}
\newcommand{\pred}{\text{known}}
\newcommand{\calib}{\text{calib}}
\newcommand{\train}{\text{train}}

\newcommand{\ensembleSet}{E}
\newcommand{\ensemble}{\hat{\dynamics}}
\newcommand{\ensemblek}{\hat{\dynamics}^{k}_{\latent}}
\newcommand{\maxEnsemble}{K}
\newcommand{\disagreement}{\operatorname{D}}
\newcommand{\OODpenalty}{\kappa}

\newcommand{\alphaQuantile}{\alpha_\text{trans}}
\newcommand{\alphaCalibration}{\alpha_\text{cal}}
\newcommand{\trajectory}{\tau}
\newcommand{\threshold}{\epsilon}
\newcommand{\thresholdCalib}{\hat{\epsilon}}

\newcommand{\latentUncertainty}{\tilde{z}}

\newcommand{\uncertainty}{u}
\newcommand{\dynamicsUncertainty}{f_{\latentUncertainty}}
\newcommand{\marginfuncUncertainty}{\ell_{\latentUncertainty}}
\newcommand{\valfuncUncertainty}{\monitor}
\newcommand{\unsafeSetUncertainty}{\tilde{\mathcal{U}}} 

\newcommand{\oursname}{\textbf{\textit{UNISafe}}\xspace}
\newcommand{\ours}{\textcolor{vibrant_orange}{\oursname}\xspace}

\newcommand{\baselineLatent}{\textcolor{dark_ocean}{\textbf{\textit{LatentSafe}}}\xspace}
\newcommand{\safeOnly}{\textcolor{wine}{\textbf{\textit{SafeOnly}}}\xspace}

\newcommand{\maxVar}{\textit{MaxAleatoric}\xspace}
\newcommand{\unitVar}{\textit{TotalUncertainty}\xspace}
\newcommand{\density}{\textit{DensityEst}\xspace}

  [2023/05/06 v0.2.2 easily annotate equations using TikZ]

\RequirePackage{ifluatex}
\ifluatex
\RequirePackage{luatex85}
\RequirePackage{pdftexcmds}
  \makeatletter
  \let\pdfstrcmp\pdf@strcmp
  \let\pdffilemoddate\pdf@filemoddate
  \makeatother
\fi

\RequirePackage{tikz}
\RequirePackage{xcolor}

\usetikzlibrary{backgrounds}
\usetikzlibrary{arrows,shapes}
\usetikzlibrary{tikzmark} 
\usetikzlibrary{calc} 

\renewcommand{\eqnhighlightheight}{\mathstrut} 

\renewcommand{\eqnhighlightshade}{17}  

\renewcommand{\eqnannotationstrut}{\strut} 

\providecommand\EAmarkanchor{north} 
\providecommand\EAwesteast{east} 
\providecommand\EAlabelanchor{south} 
\pgfkeys{
    /eqnannotate/.is family, /eqnannotate,
    above/.code = {\renewcommand\EAmarkanchor{north}},
    below/.code = {\renewcommand\EAmarkanchor{south}},
    left/.code = {\renewcommand\EAwesteast{west}},
    right/.code = {\renewcommand\EAwesteast{east}},
    label above/.code = {\renewcommand\EAlabelanchor{south}},
    label below/.code = {\renewcommand\EAlabelanchor{north}},
}

\tikzset{annotate equations/arrow/.style={}}
\tikzset{annotate equations/text/.style={font=\eqnannotationfont}}

\renewcommand*{\eqnhighlightcolorbox}[2]{%
    \mathchoice
        {\colorbox{#1}{$\displaystyle #2$}}%
        {\colorbox{#1}{$\textstyle #2$}}%
        {\colorbox{#1}{$\scriptstyle #2$}}%
        {\colorbox{#1}{$\scriptscriptstyle #2$}}%
}

\renewcommand*{\eqnhighlightfbox}[2]{%
    \mathchoice
        {\begingroup\setlength{\fboxrule}{0pt}\fbox{$\displaystyle\color{#1}#2$}\endgroup}%
        {\begingroup\setlength{\fboxrule}{0pt}\fbox{$\textstyle\color{#1}#2$}\endgroup}%
        {\begingroup\setlength{\fboxrule}{0pt}\fbox{$\scriptstyle\color{#1}#2$}\endgroup}%
        {\begingroup\setlength{\fboxrule}{0pt}\fbox{$\scriptscriptstyle\color{#1}#2$}\endgroup}%
}

\renewcommand*{\eqnhighlight}[2]{\begingroup\colorlet{currentcolor}{.}\eqnhighlightcolorbox{#1!\eqnhighlightshade}{\eqnhighlightheight #2}\endgroup}
\renewcommand*{\eqncolor}[2]{\begingroup\colorlet{currentcolor}{.}\eqnhighlightfbox{#1}{\eqnhighlightheight #2}\endgroup}

\renewcommand*{\eqnmarkbox}[3][currentcolor]{\addvalue{#2}{#1}\tikzmarknode{#2}{\eqnhighlight{#1}{#3}}}

\def\addvalue#1#2{\expandafter\gdef\csname eqnannotate@data@#1\endcsname{#2}}
\def\usevalue#1{%
  \ifcsname eqnannotate@data@#1\endcsname
    \csname eqnannotate@data@#1\expandafter\endcsname
  \else
    currentcolor%
  \fi
}

\renewcommand*{\swapNorthSouth}[1]{%
    \ifnum\pdfstrcmp{#1}{south}=0 north\else south\fi
}
\renewcommand*{\swapWestEast}[1]{%
    \ifnum\pdfstrcmp{#1}{east}=0 west\else east\fi
}
\renewcommand*{\EAxshift}[1]{%
    \ifnum\pdfstrcmp{#1}{east}=0 -0.3ex\else 0.3ex\fi
}

\renewcommand*{\eqnannotateCurrentNode}{eqnannotatenode\theeqnannotatenode}

\RequirePackage{expl3}
\RequirePackage{xparse}
\ExplSyntaxOn
\RenewDocumentCommand{\extractfirst}{mm}
 {
  \tl_set:Nx #1 {\clist_item:Nn #2 { 1 } }
 }
\ExplSyntaxOff

\renewcommand{\annotate}[4][]{%
    \begingroup
    \stepcounter{eqnannotatenode}%
    \pgfkeys{/eqnannotate, #2}%
    \def\myEAmarks{#3}%
    \extractfirst\myEAmark\myEAmarks
    \def\myEAtext{#4}%
    \colorlet{currentcolor}{.}%
    \def\myEAcolor{\usevalue{\myEAmark}}%
    \def\EAspace{ }
    \edef\myEAlabelanchor{\EAlabelanchor\EAspace\EAwesteast}%
    \def\myEAxshift{\EAxshift{\EAwesteast}}%
    \begin{tikzpicture}[overlay,remember picture,>=stealth,nodes={align=left,inner ysep=1pt},<-]
		\node[anchor=\swapNorthSouth{\EAmarkanchor} \swapWestEast{\EAwesteast},
			color=\myEAcolor!85,annotate equations/text,#1] 
            (\eqnannotateCurrentNode) at (\myEAmark.\EAmarkanchor)  
            {\myEAtext\eqnannotationstrut};
        \foreach \EAmark in \myEAmarks
        \draw [color=\myEAcolor, annotate equations/arrow] (\EAmark.\EAmarkanchor)  
            |- ([xshift=\myEAxshift,yshift=0.1ex] \eqnannotateCurrentNode.\myEAlabelanchor);
    \end{tikzpicture}%
    \endgroup
}

\title{Uncertainty-aware Latent Safety Filters \\for Avoiding Out-of-Distribution Failures}

\author{
  Junwon Seo, Kensuke Nakamura, Andrea Bajcsy\\
  Carnegie Mellon University \\
  \texttt{\{junwonse, kensuken, abajcsy\}@andrew.cmu.edu}
}

\begin{document}
\maketitle

\let\oldaddcontentsline\addcontentsline
\renewcommand{\addcontentsline}[3]{}

\begin{abstract}
Recent advances in generative world models have enabled classical safe control methods, such as Hamilton-Jacobi (HJ) reachability, to generalize to complex robotic systems operating directly from high-dimensional sensor observations. However, obtaining comprehensive coverage of all safety-critical scenarios during world model training is extremely challenging. As a result, latent safety filters built on top of these models may miss novel hazards and even fail to prevent known ones, overconfidently misclassifying risky out-of-distribution (OOD) situations as safe. To address this, we introduce an uncertainty-aware latent safety filter that proactively steers robots away from both known and unseen failures. Our key idea is to use the world model’s epistemic uncertainty as a proxy for identifying unseen potential hazards. We propose a principled method to detect OOD world model predictions by calibrating an uncertainty threshold via conformal prediction. By performing reachability analysis in an augmented state space--spanning both the latent representation and the epistemic uncertainty--we synthesize a latent safety filter that can reliably safeguard arbitrary policies from both known and unseen safety hazards. In simulation and hardware experiments on vision-based control tasks with a Franka manipulator, we show that our uncertainty-aware safety filter preemptively detects potential unsafe scenarios and reliably proposes safe, in-distribution actions. Video results can be found on the project website: \hyperref[https://cmu-intentlab.github.io/UNISafe]{\textcolor{vibrant_orange}{https://cmu-intentlab.github.io/UNISafe}}

\end{abstract}



\section{Introduction}
\begin{figure*}[t!]
\centering
\includegraphics[width=1.0\linewidth]{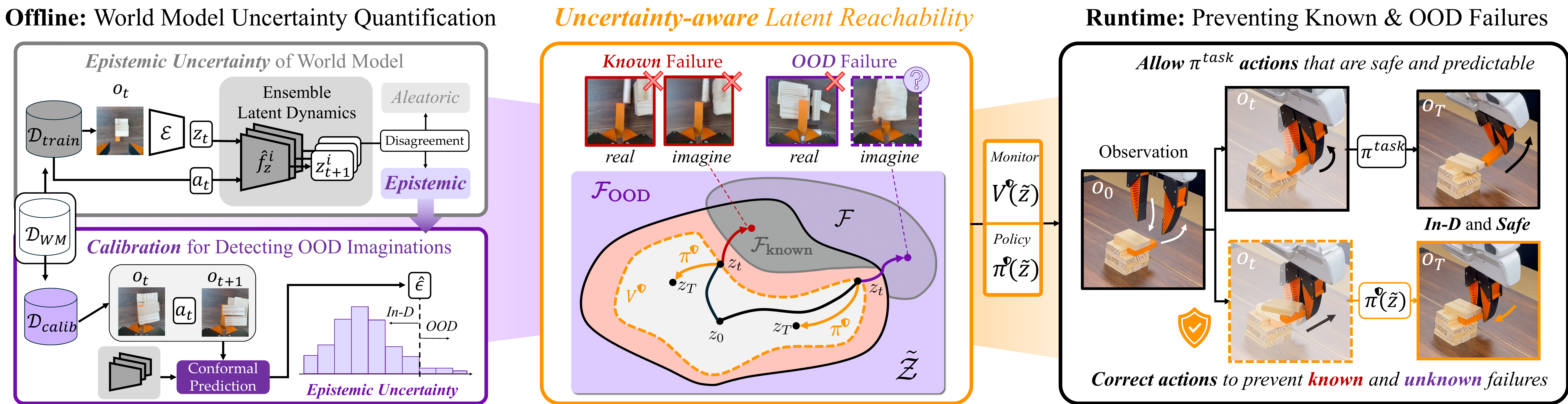}
\caption{\small{\textit{Left}: We quantify the world model’s epistemic uncertainty for detecting unseen failures in latent space and calibrate an uncertainty threshold via conformal prediction, resulting in an \textit{OOD failure set}, \(\failure_\ood\). \textit {Center}: Uncertainty-aware latent reachability analysis synthesizes a safety monitor \(\monitor\) and fallback policy \(\fallback\) that steers the system away from both known and OOD failures. \textit{Right}: Our safety filter reliably safeguards arbitrary task policies during hard-to-model vision-based tasks, like a teleoperator playing the game of Jenga.}
}
\label{fig:main}
\vspace{-0.2in}
\end{figure*}

Robots operating in complex open-world environments must interact safely with the world based on high-dimensional sensor observations. A promising approach to scale safe control to such settings is to learn a world model~(WM)~\cite{hafner2019learning} that jointly compresses observations into compact latent representations and predicts their dynamics, allowing the robot to anticipate the consequences of candidate actions to prevent unsafe ones~\cite{nakamura2025generalizing}. 
However, without unlimited unsafe exploration, the WM's training data can fail to capture the full range of possible safety hazards. 
For example, in the \textit{Jenga} game (right, Fig.~\ref{fig:main}), most of the ways in which the tower can fall are not seen during training. During interaction, if the robot fails to reliably predict how its actions can lead to such \textit{out-of-distribution (OOD)} scenarios, it may inadvertently execute actions that lead to unsafe outcomes~\cite{AgiaSinhaEtAl2024, xu2025can}. 

One way to address this model uncertainty is through OOD detection, which identifies when the robot encounters anomalous observations or generates uncertain predictions~\cite{sinha2022system, salehi2022a, yang2024generalized}. However, on its own, OOD detection lacks actionable mitigation strategies, leaving robots \textit{aware} of their uncertainty yet unable to \textit{act} appropriately. Here, safe control methods such as Hamilton-Jacobi (HJ) reachability analysis~\cite{mitchell2005time, wabersich2023data} offer a complementary approach by synthesizing fallback policies that proactively enforce safety constraints, keeping the system within control-invariant sets. Yet, they typically assume a perfect state representation and a faithful dynamics model, assumptions that may not hold in OOD scenarios when relying on a world model for safe control. 
To bridge this gap, we argue that safety constraints for latent-space control should be augmented to identify unreliable model predictions, enabling the synthesis of a \textit{safety filter} that prevents the system from entering both \textit{known failures} and potentially unsafe \textit{OOD failures}.

In this work, we propose \textbf{\textit{UNcertainty-aware Imagination for Safety filtering} (\oursname)}: a policy-agnostic safety mechanism that reliably steers robots away from known and unseen safety hazards using a latent world model~\cite{hafner2019learning, hafner2023dreamerv3}. Our key idea is to use the world model’s epistemic uncertainty as a proxy for identifying unseen potential hazards. We propose a principled method to quantify the epistemic uncertainty of the world model and detect unreliable world model predictions by calibrating an uncertainty threshold via conformal prediction. By performing reachability analysis in an augmented state space spanning both the latent states and the uncertainty, we synthesize a safety filter that can reliably prevent a system from entering both predictable and unforeseen failure modes.

We evaluate our framework in simulation and hardware on three vision-based safe-control tasks. We find that \oursname effectively prevents failures with world models trained on an offline dataset with limited coverage. Importantly, by penalizing overly optimistic safety evaluations of OOD scenarios during reachability analysis, our safety filter preemptively detects potential safety risks and proposes reliable backup actions, consistently guiding the system toward safe, in-distribution behaviors.

\section{Related Works}
\para{Out-of-distribution Detection for Robotics} Data-driven control often exhibits unreliable behavior when encountering data that deviates from its training distribution~\cite{salehi2022a, sinha2022system, AgiaSinhaEtAl2024, Sinha-RSS-24, xu2025can}.  To detect such out-of-distribution (OOD) conditions, uncertainty is estimated via pre-trained feature spaces~\cite{wong2022error, liu2024model}, reconstruction~\cite{Richter2017SafeVN, wellhausen2020safe, ruff2021unifying, schmid2022self}, density estimation~\cite{durkan2019neural, liu2020energy, kang2022lyapunov, reichlin2022back, castaneda2023distribution}, or ensembles~\cite{kidambi2020morel, yu2020mopo, rafailov2021offline, seyde2022learning}. While these methods can detect OOD and serve as runtime monitors~\cite{chua2018deep, sekar2020planning, mendonca2021discovering, AgiaSinhaEtAl2024}, they often lack control invariance, limiting them to passive detection rather than proactive failure prevention. Moreover, they typically do not distinguish between epistemic uncertainty (i.e., lack of knowledge) and aleatoric uncertainty (i.e., inherent noise)~\cite{kidambi2020morel, yu2020mopo, rafailov2021offline}, while capturing epistemic uncertainty is critical for reliable OOD detection~\cite{shyam2019model, kim2023bridging}. To bridge this gap, we quantify the epistemic uncertainty of a world model~\cite{shyam2019model, vlastelica2021riskaverse, kim2023bridging} to formulate a constraint, enabling reachability analysis to synthesize control strategies that prevent the system from OOD scenarios.

\para{Safety Filtering} Safety filtering is a control-theoretic approach for safeguarding robotic systems from unsafe conditions~\cite{mitchell2005time, ames2016control, fisac2018general, wabersich2023data, hsu2023safety, ganai2024hamilton}. While they can provide robust safety assurances under model uncertainty~\cite{fisac2018general, herbert2021scalable, hsu2023isaacs, wang2024providing, ciftci2024safe}, they focus on worst-case disturbances, addressing aleatoric uncertainty rather than epistemic uncertainty of the model. Self-supervised~\cite{bansal2021deepreach} and reinforcement learning methods~\cite{fisac2019bridging, hsu2021safety} have been used to scale safety filtering to high-dimensional systems, but these approaches typically rely on known system dynamics with simple safety specifications~\cite{hsu2023isaacs, chakraborty2024enhancing} or online rollouts in simulators~\cite{hsu2023sim, AgileButSafe, nguyen2024gameplay}. To generalize safety filters with complex dynamics and constraints, latent world models~\cite{hafner2019learning} have been used~\cite {castaneda2023distribution, wilcox2022ls3, nakamura2025generalizing}, but the epistemic uncertainty of the learned model can compromise reliability~\cite{rafailov2021offline}. Recent works prevent the system from entering OOD states~\cite{kang2022lyapunov, castaneda2023distribution}, but they restrict in-distribution to safe trajectories or do not construct constraints with calibrated OOD detection~\cite{robey2020learning, castaneda2023distribution, lindemann2024learning}, limiting their scalability to complex settings. Our method leverages calibrated OOD detection, enabling reliable prevention of both known and unseen failures.

\section{Setup: Latent Safety Filters via Reachability Analysis in a World Model} \label{sec:pre}

\vspace{-0.05in}
In this section, we briefly introduce the computation and use of latent safety filters~\cite{nakamura2025generalizing} for systems with hard-to-model dynamics and safety specifications inferred from high-dimensional observations.

\vspace{-0.05in}
\para{Latent World Model} To model complex systems, we train a world model~\cite{hafner2019learning} using a fixed offline dataset of robot–environment interactions, \(\dataset_\train := \{\{(\obs_t, \action_t, \latentfailuremargin_t)\}_{t=1}^T\}_{i=1}^{N_\train} \subset \dataset_\indist\), consisting of trajectories with high-dimensional observations \(\obs \in \obsSpace\), robot actions \(\action \in \actionSpace\), and failure labels \(\latentfailuremargin \in \{-1, 1\}\) indicating visible safety hazards. The latent world model consists of an encoder $\encoder$ that maps an observation into the latent representation $\latent \in \latentSpace$ and a latent dynamics model:
\begin{equation}
        \text{Encoder: } \latent_t \sim \encoder(\latent_t \mid \hat{\latent}_t, \obs_t)  \quad \text{Dynamics: } \hat{\latent}_t \sim \dynz(\hat{\latent}_t \mid \latent_{t-1}, \action_{t-1}) \quad \text{Failure: } \latentfailuremargin_t = \ellz(\latent_t).  \label{eq:world_model}
\end{equation}

\vspace{-0.1in}
\para{Safety Specification ($\failure$)} Hard-to-model safety constraints (e.g., spilling, block toppling) are specified in the latent space via a failure set $\failure := \{\latent : \ellz(z) \leq 0 \} \subset \latentSpace$ encoded via the zero-sublevel set of a margin function $\ellz$ in \eqref{eq:world_model}. In practice, $\ellz$ is a binary classifier learned with $\dataset_\train$.

\vspace{-0.05in}
\para{Computing Latent Safety Filters ($\fallback, \valfuncUncertainty$)} Following~\cite{nakamura2025generalizing}, we conduct HJ reachability analysis~\cite{mitchell2005time, hsu2023safety} in the latent space to synthesize both a safety value function \(\monitor: \latentSpace \rightarrow \mathbb{R}\) and a safety-preserving policy \(\fallback: \latentSpace \rightarrow \actionSpace\), entirely within the imagination of the world model. 
Specifically, we solve the fixed-point safety Bellman equation with a time discounting factor $\gamma \in [0,1)$~\cite{fisac2019bridging}:
\begin{equation}\label{eqn:discounted-safety-bellman_uncertainty}
    \valfuncUncertainty(\latent_t) = (1-\gamma)\ellz(\latent_t) + \gamma \min \Big\{ \ellz(\latent_t), \max_{\action_t \in \actionSpace} \valfuncUncertainty\big( \hat{\latent}_{t+1}\big) \Big\}, \quad \fallback(\latent_t) = \arg\max_{\action \in \actionSpace} \valfuncUncertainty( \hat{\latent}_{t+1}),
\end{equation} where $\hat{\latent}_{t+1}$ is sampled from $\dynz$. Intuitively, \(\valfuncUncertainty\) represents how close the robot comes to failure starting from \(\latent_t\) despite its best efforts, and  \(\fallback\) is a maximally safety-preserving policy. Note that, in contrast to typical RL for reward maximization, this optimization performs a \textit{min-over-time} to \textit{remember} safety-critical events. Therefore, \(\valfuncUncertainty < 0\) indicates that the robot is doomed to fail, while \(\valfuncUncertainty \geq 0\) means that there exists a safety-preserving action to prevent failures (e.g., returned by \(\fallback\)).

\vspace{-0.05in}
\para{Runtime Safety Filtering} At runtime, the latent safety filter safeguards an arbitrary task policy \(\policyTask\) based on the current observations and proposed action. By checking \(\monitor\) as a monitor with a small margin $\delta \approx 0$, the safety filter either allows \(\policyTask\) or overrides it with the fallback policy \(\fallback\):
\begin{equation}\label{eq:latent-safe-control}
    \action^\text{exec} := \mathds{1}\{\eqnmarkbox[dark_green]{safe}{\monitor(\latent')> \delta}\} \, \policyTask + \mathds{1}\{\eqnmarkbox[red]{unsafe}{\monitor(\latent') \le \delta}\} \, \fallback(\latent), \quad \latent' \sim \dynz(\latent, \policyTask).
\end{equation}
\annotate[yshift=-0.2em]{below}{safe}{$\policyTask$ is safe, proceed}
\annotate[yshift=-0.2em]{below}{unsafe}{$\policyTask$ is unsafe, fallback to $\fallback$}

\paragraph{\textit{Challenge: Unreliable WM Can Result in OOD Failures.}} While latent safety filters can compute control strategies that prevent hard-to-model failures, their training~\eqref{eqn:discounted-safety-bellman_uncertainty} and runtime filtering \eqref{eq:latent-safe-control} rely on imagined futures generated by the latent dynamics model. However, a pretrained world model can hallucinate in uncertain scenarios where it lacks knowledge, leading to \textit{OOD failures}. \begin{wrapfigure}{l}{0.54\textwidth}
    \centering
    \vspace{-0.1in} 
    \includegraphics[width=\linewidth]{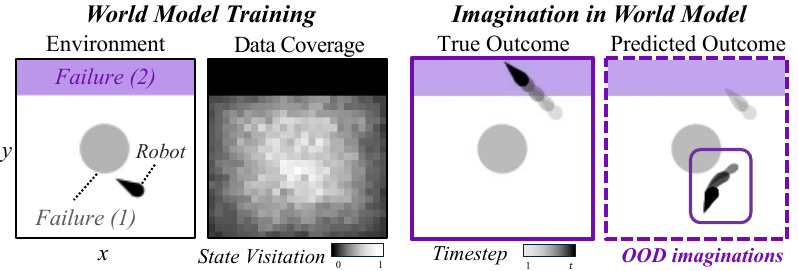}
    \vspace{-0.22in}
    \caption{\small{WM imaginations can lead to \textbf{\textit{\textcolor{violet}{OOD Failures}}}.}}
    \vspace{-0.12in}
    \label{fig:ood_running_example}
\end{wrapfigure} Consider the simple example in Fig.~\ref{fig:ood_running_example} where a Dubins car must avoid two failure sets: a circular grey and a rectangular purple region. The world model is trained with RGB images of the environment and angular velocity actions, but the model training data is limited, lacking knowledge of the robot entering the purple failure set. When the world model imagines an action sequence in which the robot enters this region (\textit{third} image of Fig.~\ref{fig:ood_running_example}), the world model hallucinates as soon as the scenario goes out-of-distribution: the robot teleports away from the failure region and to a safe state (\textit{rightmost} image of Fig.~\ref{fig:ood_running_example}). This phenomenon leads to latent safety filters that cannot prevent unseen failures, and even known failures, due to optimistic safety estimates of uncertain out-of-distribution scenarios.

\section{Formalizing Uncertainty-aware Latent Safety Filters}

To formalize reliable safe control in latent space, our key idea is to use the epistemic uncertainty of the world model as a proxy for detecting safety hazards not represented in the training dataset. 
Specifically, we augment the safety specification that accounts for \textbf{\textit{known failures}}--scenarios the world model can anticipate with confidence--with \textbf{\textit{OOD failures}}: potentially unsafe, out-of-distribution scenarios where the model's imaginations are highly uncertain and lose their reliability.

\vspace{-0.1in}
\paragraph{Uncertainty-aware Latent Space \& Dynamics.} We quantify the epistemic uncertainty of the world model, \(\uncertainty \in \mathbb{R}\), to identify OOD imaginations of the world model. To assess the reliability of latent dynamics predictions, the uncertainty should capture the \textit{dynamics uncertainty} induced by latent–action transitions \((\latent, \action)\). This is crucial because generative world models are prone to hallucination, often producing in-distribution predictions when exposed to OOD inputs. Therefore, OOD detection methods that rely solely on the predicted latent state \(\latent\) are overconfident, as predicted latents from OOD scenarios are projected into in-distribution representations, as depicted in Fig.~\ref{fig:ood_running_example}.

We then augment the latent space to incorporate this epistemic uncertainty, $\latentUncertainty_t = (\latent_t, \uncertainty_t)^\top \in \latentSpace \times \mathbb{R}$. This formulation enables modeling both \textit{known failures} \(\failure_\pred := \{ \latentUncertainty \mid \ellz(\latent) < 0 \}\), which are predictable with the learned model, and \textit{OOD failures} \(\failure_\ood := \{ \latentUncertainty \mid \uncertainty > \threshold \}\) which are OOD imaginations with quantified uncertainty exceeding a predefined threshold~\(\threshold\). The latent dynamics and safety margin function are extended to operate in the augmented latent space:
\begin{equation}\label{eq:uncertainty-aware-latent}
\dynamicsUncertainty(\latentUncertainty_{t+1} \mid \latentUncertainty_t, \action_t) = \big[ \dynz(\latent_{t+1} \mid \latent_t, \action_t), ~\disagreement(\latent_t, \action_t) \big]^\top, \quad
\marginfuncUncertainty(\latentUncertainty_t) = \min\big\{ \ellz(\latent_t)\,,  \OODpenalty \, (\threshold - \uncertainty_t) \big\},
\end{equation} with \(\OODpenalty \in \mathbb{R}^{+}\). The uncertainty \(\uncertainty_{t+1} = \disagreement(\latent_{t}, \action_{t})\) is obtained via measuring reliability of a transition (described in Sec~\ref{sec:uq-wm}) and the uncertainty-aware failure set $\failureTotal$ is represented via the zero sub-level set of the augmented margin function: 
$\failureTotal := \failure_\pred \cup \failure_\ood = \{ \latentUncertainty \; | \; \marginfuncUncertainty(\latentUncertainty) < 0 \}$.

\vspace{-0.1in}
\paragraph{Uncertainty-aware Latent Reachability Analysis.} We compute a latent safety filter via Eq.~\ref{eqn:discounted-safety-bellman_uncertainty} and perform safety filtering as in Eq.~\ref{eq:latent-safe-control}, but use the uncertainty-aware latent dynamics from Eq.~\ref{eq:uncertainty-aware-latent} throughout. This formulation ensures that the value function assigns negative values to OOD scenarios where the uncertainty exceeds a predefined threshold. By explicitly penalizing such transitions, the resulting safety filter discourages the system from entering OOD regions while also avoiding known, predictable failures. This mitigates overly optimistic imaginations and enables the filter to reliably learn both a safety monitor and a fallback policy that proposes safe, in-distribution actions.

\section{Computing Uncertainty-aware Latent Safety Filters}
While the prior section formalize the uncertainty-aware latent safety filter by augmenting the latent space with the world model’s epistemic uncertainty, we face two key challenges when instantiating our framework in practice: (\romannumeral 1) How can we quantify the epistemic uncertainty of the world model? (Sec.~\ref{sec:uq-wm}) and (\romannumeral 2) How can we ensure the OOD threshold $\threshold$ is appropriately \textit{calibrated} to reliably detect OOD failures based on the estimated measures of epistemic uncertainty? (Sec.~\ref{sec:wm-calib}).

\subsection{Quantifying Epistemic Uncertainty in World Models}\label{sec:uq-wm}
\paragraph{Training a Probabilistic Ensemble Latent Predictor.}  
To capture the epistemic uncertainty of the world model, we employ an ensemble of next-latent predictors, \(\ensembleSet := \{\ensemblek\}_{k=1}^{\maxEnsemble}\), which is a separate module regressing the pretrained latent dynamics \(\dynz\). Each ensemble member is initialized with distinct parameters \(\ensembleparam_k\) and trained to predict the next latent \(\latent_{t+1}\) given the current latent \(\latent_t\) and action \(\action_t\) with Gaussian negative log-likelihood loss (see \ref{sec:appendix_ensemble} \& \ref{sec:ensembleRSSM} for more details.):
\begin{equation}
\text{Latent Predictor:} \quad \latent_{t+1}^k \sim \ensemblek(\latent_t, \action_t; \ensembleparam_k), \quad \ensemblek(\latent_t, \action_t; \ensembleparam_k) := \mathcal{N}(\mu_{\ensembleparam_k}(\latent_t, \action_t), \Sigma_{\ensembleparam_k}(\latent_t, \action_t)),
\label{eq:ensemble}
\end{equation} where \(\mu_{\ensembleparam_k}\) and \(\Sigma_{\ensembleparam_k}\) denote the predicted mean and diagonal covariance, respectively. Note that the covariance models the \textit{aleatoric uncertainty} inherent in the latent dynamics due to partial observability and stochasticity. The ensemble latent predictor is trained on latent transitions \(\{\{(\latent_t, \action_t, \latent_{t+1})\}_{t=1}^{T-1}\}_{i=1}^{N_\train}\), encoded from a pretrained latent world model with the same offline dataset \(\dataset_\train\) used for world model training.

\paragraph{Epistemic Uncertainty Quantification.}~\label{sec:epistemic-uncertainty-quantificaiton} While empirical variance over ensemble predictions is widely used as an uncertainty measure~\cite{kidambi2020morel, yu2020mopo, sekar2020planning, wang2024providing}, this conflates aleatoric uncertainty (i.e., inherent uncertainty of latent dynamics) with epistemic uncertainty (i.e., uncertainty arising from a lack of knowledge). Since our goal is to control away from \textit{OOD failures}, which the world model has never encountered and thus cannot reliably predict, it is essential to focus explicitly on the model’s \textit{epistemic uncertainty} to form our constraint on the latent space. Otherwise, the safety filter may fail to reject unsafe OOD imaginations or become overly conservative in response to intrinsic stochasticity. Following~\cite{shyam2019model, kim2023bridging}, we quantify the epistemic uncertainty of the latent dynamics $\disagreement(z_t,a_t)$ via the Jensen-Rényi Divergence~(JRD)~\cite{renyi1961measures} of the ensemble predictions with Rényi entropy $H_\alpha$:\begin{equation}
    \eqnmarkbox[grape]{epistemic}{\uncertainty_{t+1} = \disagreement(z_t,a_t)} :=  \eqnmarkbox[black]{total}{H_{\alpha}\left(\sum_{k=1}^{K} \frac{1}{K} \ensemblek \right)} - \eqnmarkbox[grey]{aleatoric}{\sum_{k=1}^{\maxEnsemble} \frac{1}{K} H_{\alpha}\left(\ensemblek\right)}, \quad  H_{\alpha}(Z)=\frac{1}{1-\alpha} \log \int p(z)^{\alpha} \mathrm{d}z,
\end{equation} 
\annotate[yshift=-1.2em]{below}{epistemic}{epistemic uncertainty}
\annotate[yshift=-0.2em]{below}{total}{total uncertainty}
\annotate[yshift=-0.2em]{below}{aleatoric}{aleatoric uncertainty}
\vspace{-0.1in}

where $Z$ is a random variable. In general, computing the disagreement between an ensemble of Gaussian distributions lacks a closed-form solution~\cite{shyam2019model, kim2023bridging}, and Monte Carlo sampling approximations are computationally expensive for high-dimensional latent spaces. Therefore, we adopt JRD with $\alpha=2$, which has a closed-form expression~\cite{wang2009closed} for GMM (see \ref{sec:appendix_ensemble} for further details).

\subsection{Detecting Out-of-Distribution Imaginations via Conformal Prediction}\label{sec:wm-calib}
Recall that during reachability analysis, \textit{OOD failures} are detected when the uncertainty of an imagined transition exceeds a threshold, \(\failure_\ood = \{ \latentUncertainty \mid \uncertainty > \threshold \}\). 
However, setting this threshold is nontrivial: too strict a threshold can result in high false-positive rates (misclassifying in-distribution transitions as OOD), leading to overly conservative filters; too loose a threshold may fail to detect true OOD transitions. We employ conformal prediction (CP)~\cite{vovk2005algorithmic, angelopoulos2023conformal} to automatically calibrate the threshold \(\threshold \in \mathbb{R}\) in a principled way, using a held-out calibration dataset \(\dataset_\calib = \dataset_\indist \setminus \dataset_\train\). 

\vspace{-0.1in}
\paragraph{In-distribution Recall Guarantee via Class-Conditioned Conformal Prediction.} 
CP typically requires the calibration set $\dataset_\calib$ to contain both inputs to the prediction model (e.g., ($\latent_t, \action_t$)) and their corresponding ground-truth labels (e.g., ID or OOD). Unfortunately, in our setting, true OOD labels are, by definition, not accessible. As such, we assume the calibration dataset consists only of in-distribution transitions.  Formally, we adopt class-conditioned conformal prediction~\cite{ding2023class, chakraborty2024enhancing} to calibrate the uncertainty threshold $\threshold$, providing conditional recall guarantees for detecting in-distribution transitions with user-defined confidence level $\alphaCalibration \in [0,1]$: \begin{equation}\label{eq:class-conditioned-guarantee}
    \mathbb{P}\left( \disagreement(\latent_t, \action_t) < \thresholdCalib \mid (\latent_t, \action_t) \in \dataset_{\indist} \right) \geq 1 - \alphaCalibration,
\end{equation} Intuitively, conformal prediction can help us select an uncertainty threshold $\thresholdCalib$ such that in-distribution latent transitions can be detected with probability at least $1-\alphaCalibration$. Conversely, latent transitions with uncertainty greater than this threshold can be interpreted as OOD.

\vspace{-0.1in}
\paragraph{Trajectory-Level Calibration.} While standard class-conditioned conformal prediction assumes exchangeability of the data, this assumption does not hold in our setting, as each transition depends on the full history of latent states and actions. 
To address this, we adopt a trajectory-level calibration approach~\cite{ren2023robots}, assuming that the calibration trajectories \(\trajectory_i = \{(\latent_t, \action_t)\}_{t=1}^{T} \in \dataset_\calib\)  are drawn i.i.d. from the same distribution as the world model training data, \(\{\trajectory_i\}_{i=1}^{N} \overset{\mathrm{iid}}{\sim} \dataset_\indist\). 
For each trajectory, we define the trajectory-level nonconformity score \(Q_{\trajectory_i}^{\alphaQuantile}\) as the \((1 - \alphaQuantile)\)-quantile of the set of quantified epistemic uncertainties \(\{\uncertainty_t\}_{t=1}^{T}\). This ensures that at most an \(\alphaQuantile\) fraction of a trajectory’s uncertainty values exceed \(Q_{\trajectory_i}^{\alphaQuantile}\), making the estimate more robust to noise in uncertainty predictions. We then determine the calibration threshold \(\thresholdCalib\) as the \((1 - \alphaCalibration)\)-quantile of the set \(\{Q_{\trajectory_i}^{\alphaQuantile}\}_{i=1}^{N}\) by selecting the \(\lceil (1 - \alphaCalibration)(N + 1) \rceil\)-th smallest value over trajectories. With the exchangeability assumption between calibration and test trajectories, conformal prediction guarantees that for a new test trajectory \(\trajectory_{\text{test}} = \{(\latent^{\text{test}}_t, \action^{\text{test}}_t)\}_{t=1}^{T}\), the following probabilistic guarantee holds:
\begin{equation}\label{eq:coformal_guarantee}
  \mathbb{P}_{\trajectory_\text{test} \sim \dataset_\indist} (Q_{\trajectory_\text{test}}^{\alphaQuantile} \leq \thresholdCalib) = \mathbb{P}_{\trajectory_\text{test} \sim \dataset_\indist} \left( \mathbb{P}_{t} \left\{ \disagreement(\latent^{\text{test}}_t, \action^{\text{test}}_t) \le \thresholdCalib \right\} \ge 1 - \alphaQuantile \right) \ge 1 - \alphaCalibration.  
\end{equation} 
Although this guarantee applies only to in-distribution data, it ensures a low false positive rate by bounding the probability of misclassifying in-distribution transitions as OOD. Specifically, the probability that the trajectory-level nonconformity score exceeds the threshold for in-distribution data is bounded by \(\mathbb{P}_{\trajectory_\text{test} \sim \dataset_\indist}\left( Q_{\trajectory_{\text{test}}}^{\alphaQuantile} \ge \thresholdCalib \right) \le \alphaCalibration\). As a result, any transition with a quantified epistemic uncertainty above \(\thresholdCalib\) can be reliably classified as OOD, since such events are guaranteed to be rare under the in-distribution distribution (see Appendix~\ref{sec:appendix_conformal} for details).

\section{Simulation \& Hardware Experiments}

\subsection{Simulation: A Benchmark Safe Control Task with a 3D Dubins Car}\label{sec:dubins} 

We first conduct experiments with a low-dimensional, benchmark safe navigation task where privileged information about the state, dynamics, safe set, and safety controller is available.

\para{Privileged Dynamics: Dubins Car} Let the privileged Dubins car state be \(\state = [p_x, p_y, \theta]\), with discrete-time dynamics  \(\state_{t+1} = \state_t + \Delta t\, [v\cos(\theta_t),\, v\sin(\theta_t),\, a_t]\). We assume a fixed velocity \(v = \SI{1}{m/s}\), time step \(\Delta t = \SI{0.05}{s}\), and discrete action space \( a_t \in \mathcal{A} = \{-1.25, 0, 1.25\}\) rad/s.

\para{Evaluation \& Metrics} Given access to ground-truth dynamics, we compute the ground-truth safety value function using grid-based methods~\cite{mitchell2007toolbox}, enabling direct evaluation of the safety monitor \(\monitor)\)’s classification accuracy across all three state dimensions. To assess \(\fallback\), we roll out the learned policies from safe initial states with positive ground-truth safety values and measure the safety rate by checking whether the resulting trajectories remain safe without violating constraints.

\begin{wraptable}{r}{0.4\textwidth}
    \vspace{-0.15in}
    \scriptsize
    \centering
    \setlength{\tabcolsep}{4pt}
    \renewcommand{\arraystretch}{1.0}
    \resizebox{1.0\linewidth}{!}{
    \begin{tabular}{clccc}
        \toprule
        &\textbf{Method} & \textbf{TPR$\uparrow$} & \textbf{TNR$\uparrow$} & \textbf{B.Acc.$\uparrow$} \\
        \midrule
        \multirow{4}{*}{\rotatebox{90}{\textbf{UQ}}} &\unitVar  & 0.88 & 0.97 & 0.93 \\
        &\maxVar   & 0.78 & 0.88 & 0.83 \\
        &\density \,($\latent, \action$) & 0.98 & 0.87 & 0.92 \\
        &\density \,($\latent$) & 0.99 & 0.56 & 0.77 \\
        \midrule
        \multirow{3}{*}{\rotatebox{90}{\textbf{Calib}}} &\textit{JRD} ($\mathbf{\threshold = \thresholdCalib}$)& 0.93 & 0.95 & 0.94 \\
        & {\textit{JRD}} ($\threshold =\thresholdCalib + 0.3$) & 0.98  & 0.43 & 0.71 \\
        & {\textit{JRD}} ($\threshold =\thresholdCalib - 0.3$) & 0.85 & 0.96 & 0.90 \\
        \bottomrule
    \end{tabular}
    }
    \vspace{-0.05in}
    \caption{\small{Safety value function quality with different OOD detection methods.}}\label{tab:uq_methods_comparison_dubins}
    \vspace{-0.15in}
\end{wraptable} 
\para{Baselines} We evaluate \ours, which learns the uncertainty-aware unsafe set \(\unsafeSetUncertainty\) from the failure set \(\failureTotal=\failure_\pred \cup \failure_\ood\), against \baselineLatent~\cite{nakamura2025generalizing}, which considers only known failures, $\failure_\pred$. Also, we compare \textit{JRD} with other OOD detection baselines to assess uncertainty quantification. \unitVar compute variance of mean predictions across the ensemble without isolating aleatoric components~\cite{sekar2020planning, rafailov2021offline, seyde2022learning}. \maxVar uses the maximum predicted ensemble variance, \(\max_k \left\| \Sigma_{\ensembleparam_k}(z_t, a_t) \right\|_F\), representing aleatoric uncertainty~\cite{yu2020mopo, sun2023model}. \density employs neural spline flows~\cite{durkan2019neural, kang2022lyapunov} to compute likelihoods of \((\latent)\) or \((\latent, \action)\) for OOD detection. For every method, thresholds are calibrated with the same held-out calibration dataset and 
Double DQN (DDQN)~\cite{van2016deep} is used to train all the safety value function (See \ref{sec:appendix_implementation_details} and \ref{sec:appendix_experiments_details} for more details).

\begin{wrapfigure}{r}{0.38\textwidth}
    \centering
    \vspace{-0.25in}
    \includegraphics[width=\linewidth]{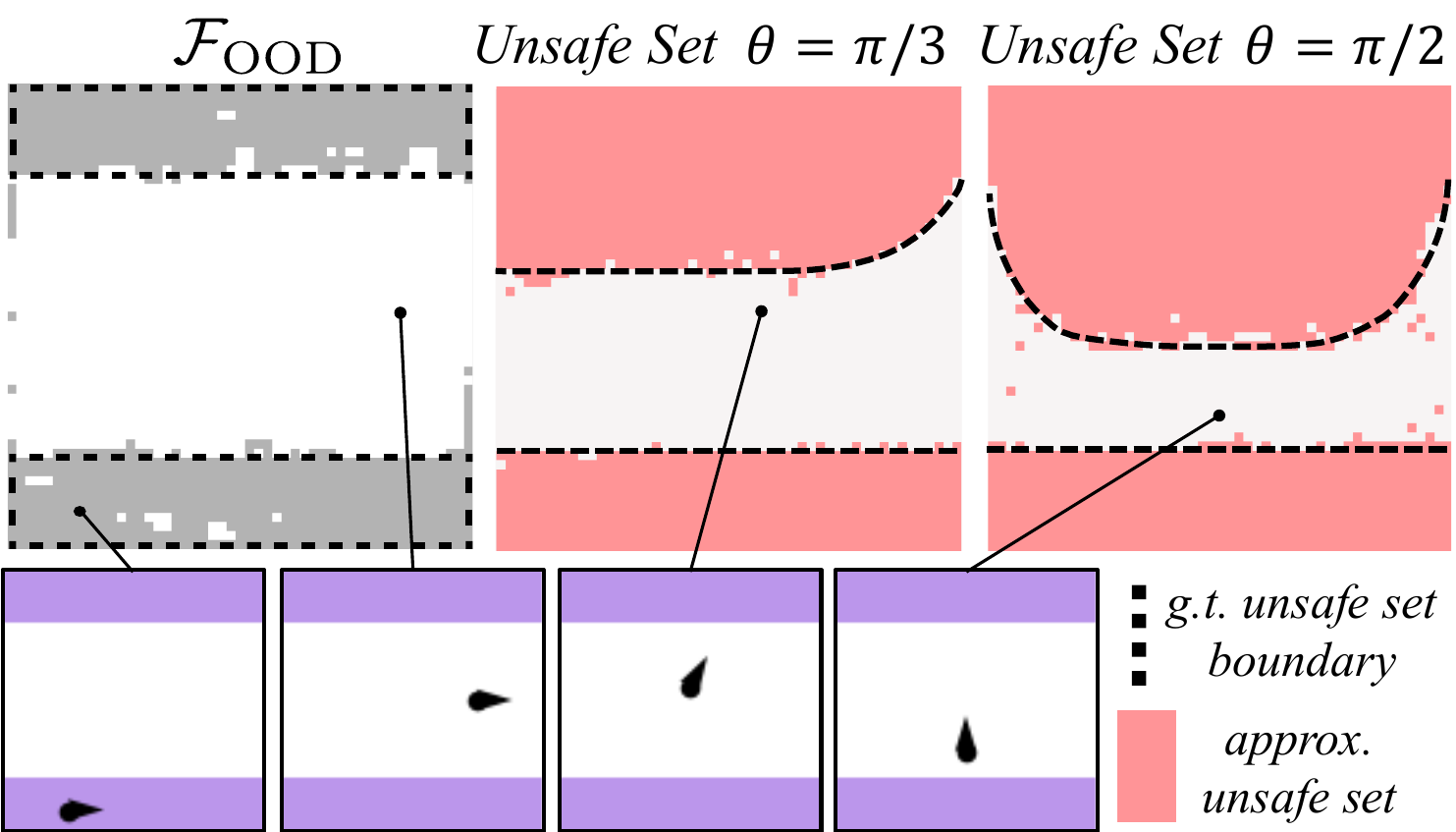}
    \vspace{-0.2in}
    \caption{\small{\textit{Dubins Car with $\failure_\ood$ only.} OOD detection successfully identifies $\failure_\ood$ and unsafe set.}}
    \label{fig:dubins_ood}
    \vspace{-0.15in}
\end{wrapfigure}
\para{\oursname reliably identifies the OOD failure $\failure_\ood$} To evaluate OOD detection, we first consider a setting where failure states are never observed by $\dataset_\train$. The ground-truth failure set is defined as \(|p_y| > 0.6\), while the offline dataset contains only $1000$ safe trajectories that never enter this region, making the failure set entirely OOD. As shown in Fig.~\ref{fig:dubins_ood}, our method reliably infers the OOD failure set from the quantified uncertainty, and the resulting safety value function accurately identifies the unsafe region. Table~\ref{tab:uq_methods_comparison_dubins} shows that \textit{JRD} achieves the highest balanced accuracy~(B.Acc.) compared to other OOD detection methods, whereas methods not targeting epistemic uncertainty exhibit higher FPRs and lower balanced accuracies. Additionally, \density based only on \(\latent\) shows low TNR, highlighting the necessity of latent-action transition-based OOD detection.

\para{A calibrated OOD threshold yields a higher quality value function} We perturb our calibrated threshold $\thresholdCalib$ to obtain $\threshold = \thresholdCalib \pm 0.3$ and study the sensitivity of the value function to threshold selection. 
Table~\ref{tab:uq_methods_comparison_dubins} shows that our automatic calibration process selects thresholds that lead to value functions with both high TPR and TNR, unlike the uncalibrated thresholds that degrade accuracy.

\para{\oursname robustly learns safety filters despite high uncertainties in the world models} We evaluate whether our method can synthesize a robust safety filter with uncertain world due to limited data coverage. In this setting, the vehicle must avoid a circular obstacle of radius \(\SI{0.5}{m}\) at center, with the failure set defined as \(p_x^2 + p_y^2  < 0.5^2\), and $\dataset_\train$ consists of both safe and unsafe trajectories. We construct a dataset of $1000$ expert trajectories that never enter the ground-truth unsafe sets and $50$ random trajectories that may include failure states.
Expert trajectories are generated using the ground-truth safety value, applying fallback actions near the unsafe boundary and random actions elsewhere, \begin{wrapfigure}{r}{0.59\textwidth}
    \centering
    \vspace{-0.2in}
    \includegraphics[width=1.0\linewidth]{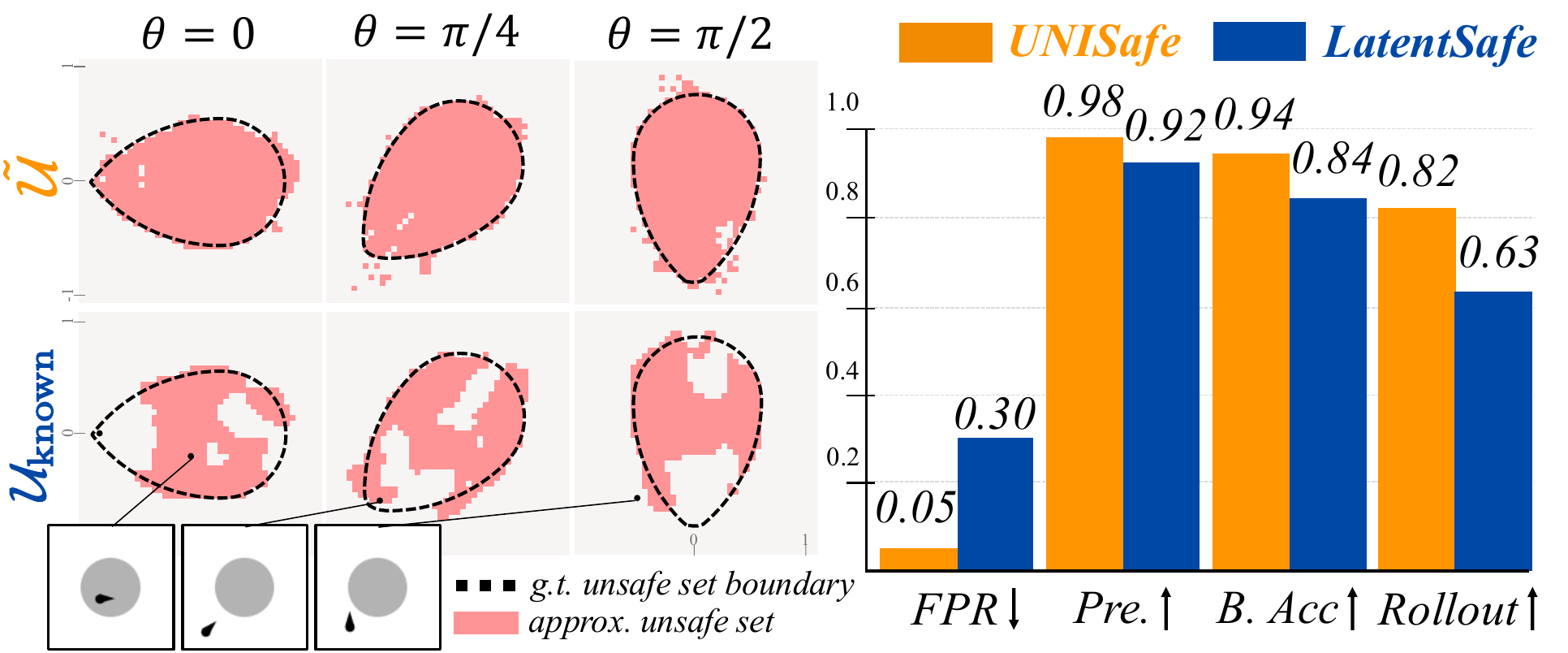}
    \vspace{-0.2in}
    \caption{\small{\textbf{\ours vs \baselineLatent}. Without \textit{OOD failures}, the safety value learned from the unreliable world model leads to higher FPR, overconfidently classifying unsafe states as safe.}}
    \label{fig:dubins_result}
     \vspace{-0.15in}
\end{wrapfigure} inducing high uncertainty around the unsafe boundary. Fig.~\ref{fig:dubins_result} shows that \ours robustly learns the safety monitor with higher balanced accuracy, whereas \baselineLatent overconfidently misclassifies unsafe states as safe. In rollouts from 181 challenging safe initial states, where the vehicle is oriented toward failure, \ours also achieves higher safety rates. (See \ref{sec:appendix_experiments_details} \& \ref{sec:dubins_ablations} for more details and analysis.)

\subsection{Simulation: Vision-Based Block Plucking}
\vspace{-0.1in}

\para{Setup} We scale our method to a visual manipulation task using IsaacLab~\cite{makoviychuk2021isaac}, where a Franka manipulator must pluck the middle block from a stack of three while ensuring the top one remains on the bottom one. Observations consist of images from a wrist-mount and a tabletop camera, with $7$-D proprioceptive inputs. Actions are a $6$-DoF end-effector delta pose with a discrete gripper command.

\vspace{-0.05in}
\begin{wrapfigure}{r}{0.48\textwidth}
    \centering
    \vspace{-0.2in}
    \includegraphics[width=1.0\linewidth]{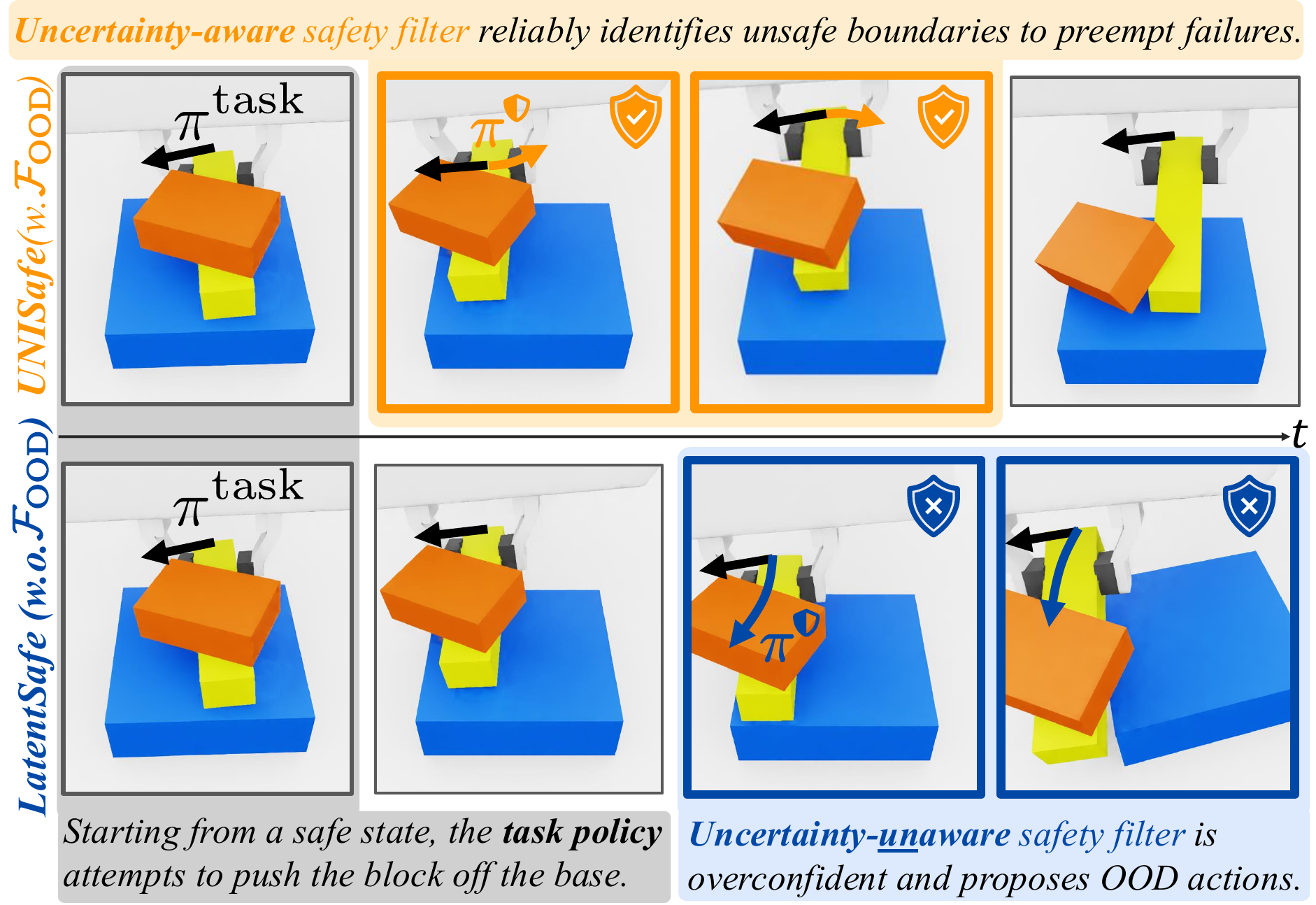}
    \vspace{-0.2in}
    \caption{\small{\textit{Block Plucking}. \ours prevents failure with ID actions, while \baselineLatent fails to preempt it by overestimating unsafe OOD actions.}}
    \label{fig:isaac_overestimation}
    \vspace{-0.2in}
\end{wrapfigure}
\para{Evaluations} We adopt DreamerV3~\cite{hafner2023dreamerv3} as our task policy \(\policyTask\), trained with a dense reward signal to achieve the task with a soft penalty for failures. The training dataset $\dataset_\train$ consists of $3000$ trajectories comprising both safe and unsafe behavior rolled out from $\policyTask$. We adopt Soft Actor-Critic (SAC)~\cite{haarnoja2018soft} as our solver for latent reachability. For evaluation, task policy rollouts are filtered using the safety filter with \(\delta = 0.1\), evaluated over $1000$ randomly sampled initial conditions. (See \ref{sec:appendix_implementation_details} and \ref{sec:appendix_experiments_details_isaac} for details.) 

\vspace{-0.07in}
\para{Baselines} As in Sec.~\ref{sec:dubins}, we compare \ours with \baselineLatent~\cite{nakamura2025generalizing} trained on the same dataset with and without $\failure_\ood$, as well as different OOD detection baselines. \safeOnly learns a WM and latent safety filter only on successful demonstrations without $\failure_\pred$, implicitly treating all failures as \(\failure_\ood\), as in~\cite{kang2022lyapunov, castaneda2023distribution, xu2025can}. Also, we adapt \textit{CQL}~\cite{kumar2020conservative} and \textit{COMBO}~\cite{yu2021combo} to optimize Eq.~\ref{eqn:discounted-safety-bellman_uncertainty} with conservative losses, but without uncertainty quantification. 

\vspace{-0.07in}
\para{\oursname minimizes failure by preventing safety overestimation} Table~\ref{tab:isaac_result} shows that \ours, which incorporates both known and OOD failures, achieves the lowest failure rates and model errors. In contrast, \baselineLatent overestimates the safety of OOD actions, leading to unsafe action proposals, as shown in Fig.~\ref{fig:isaac_overestimation}. \safeOnly shows limited effectiveness, showing OOD detection from success-only data is insufficient in complex settings. Offline RL with conservative losses performs even worse than \baselineLatent, indicating that conservatism alone cannot replace failure set identification.

\vspace{-0.05in}
\para{Quantifying epistemic uncertainty leads to safe but non-conservative behaviors} While all OOD detection methods improve filtering performance over \baselineLatent, targeting aleatoric uncertainty (\unitVar and \maxVar) tends to be overly conservative, resulting in higher incompletion rates and more frequent interventions. In contrast, \ours with \textit{JRD} explicitly targets epistemic uncertainty and achieves the most reliable performance. \density shows limited performance, highlighting the challenge of modeling likelihood in high-dimensional latent spaces.

\begin{table}[t]
    \vspace{-0.1in}
    \centering
    \scriptsize
    \setlength{\tabcolsep}{3pt}
    \renewcommand{\arraystretch}{0.7}
    \resizebox{\linewidth}{!}{
        \begin{tabular}{l|ccc|ccccc}
            \toprule
             \textbf{Method} & 
             $\dynz$ & $\failure_\pred$ & $\failure_\ood$ &
             \textbf{Safe Success ($\uparrow$)} & \textbf{Failure ($\downarrow$)} & \textbf{Incompletion} & \textbf{Filtered} ($\%$) & \textbf{Model Error ($\downarrow$)}\\ 
             \midrule
             No Filter ($\policyTask$) &
             - & - & - &
             0.58 & 0.41 & 0.01 & 0.0 ± 0.0 & 59.3 ± 3.3 \\ 
            \cmidrule(lr){1-9}
             \textit{CQL}~\cite{kumar2020conservative} & 
             \xmark & \cmark & \xmark  &
             0.63 & 0.33 & 0.04 & 2.3 ± 0.9 &  50.9 ± 11.5 \\
             \textit{COMBO}~\cite{yu2021combo} & 
             \cmark & \cmark & \xmark & 
             0.47 & 0.41 & 0.12 & 54.8 ± 6.8 & 51.6 ± 12.8 \\ 
             \cmidrule(lr){1-9}
            \safeOnly & 
            \cmark & \xmark & \cmark & 0.71 & 0.28 & 0.01 & 13.5 ± 3.5 & 46.9 ± 2.6 \\
            \baselineLatent~\cite{nakamura2025generalizing} & 
            \cmark & \cmark & \xmark & 
            0.68 & 0.30 & 0.01 & 7.2 ± 2.6 &  60.2 ± 4.7 \\ 
            \cmidrule(lr){1-9} 
            \ours (\unitVar) &
            \cmark & \cmark & \cmark & 
            0.54 & 0.18 & 0.28 & 50.9 ± 7.1 & 39.1 ± 4.1 \\
            \ours (\maxVar) & 
            \cmark & \cmark & \cmark & 
            0.64 & 0.25 & 0.11 & 38.7 ± 7.0 & 41.4 ± 9.1 \\
            \ours (\density) & 
            \cmark & \cmark & \cmark & 
            0.66 & 0.24 & 0.10 & 27.9 ± 5.1 & 41.4 ± 5.2 \\
             \ours (\textit{JRD}) & 
            \cmark & \cmark & \cmark & 
            0.72 & 0.20 & 0.08 & 37.7 ± 6.7 &  43.1 ± 1.2 \\
            \bottomrule 
        \end{tabular}
        }
    \vspace{0.05in}
    \caption{\small{\textit{Rollout Results on Block Plucking.} Safe success is plucking a block without failure, and incompletion is a timeout without success or failure. The average world model training loss per trajectory is reported as a proxy for uncertainty. Safety filter is most effective when $\dynz$ imaginations incorporate both $\failure_\pred$ and $\failure_\ood$.}}
    \label{tab:isaac_result}
    \vspace{-0.2in}
\end{table}

\subsection{Hardware: Vision-based \textit{Jenga} with a Robotic Manipulator}

\begin{figure}[h]
    \centering
    \vspace{-0.15in} \includegraphics[width=1.0\linewidth]{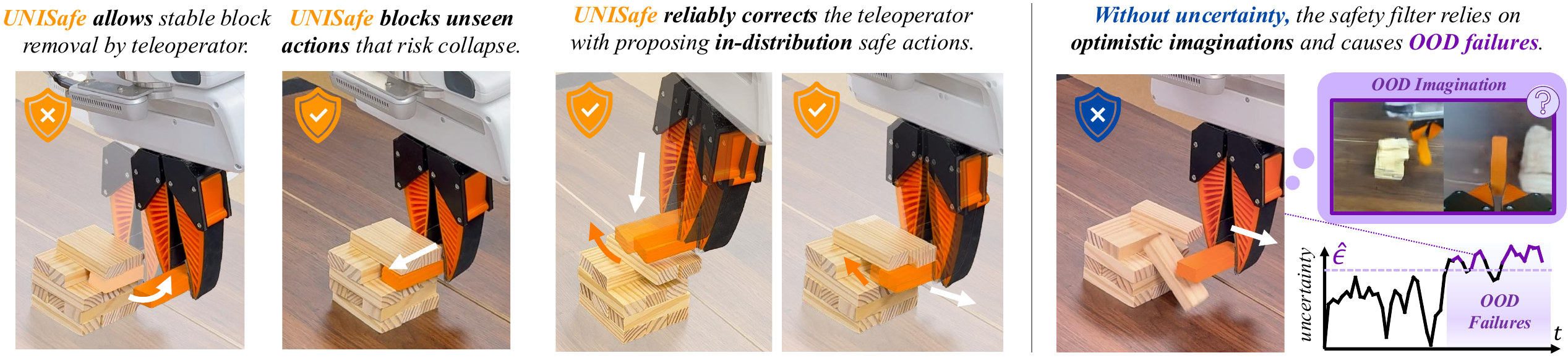}
    \vspace{-0.2in}
    \caption{\small{\textit{Teleoperator Playing Jenga with Safety Filters.} \ours enables non-conservative yet effective filtering of the teleoperator’s actions, ensuring the system remains within the in-distribution regions. In contrast, the uncertainty-unaware safety filter \baselineLatent optimistically treats uncertain actions as safe, leading to failure.} }
    \label{fig:jenga_qualitative}
    \vspace{-0.1in}
\end{figure}

\para{Setup} We evaluate our method on a real-world robotic manipulation task using a fixed-base Franka Research 3 arm, equipped with a third-person camera and a wrist-mounted camera. The robot must extract a target block from a tower without collapsing, then place it on top. For $\dataset_\train$, we collect $720$ trajectories: $150$ random (no contact), $480$ successful, and $90$ failure cases. (See \ref{sec:appendix_experiments_details_jenga} for details.)

\begin{wrapfigure}{r}{0.2\textwidth}
    \centering
    \vspace{-0.2in}
    \includegraphics[width=1.0\linewidth]{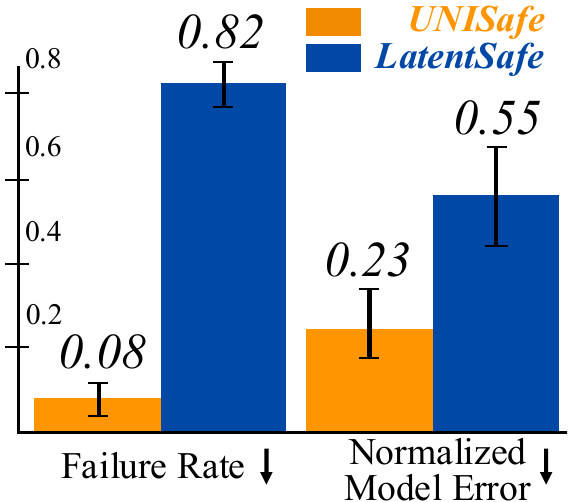}
    \vspace{-0.2in}
    \caption{\small{Filtering $\policyTask$ on hardware.}}
    \label{fig:jenga:quantitative}
    \vspace{-0.2in}
\end{wrapfigure} 

\para{\oursname reliably filters both known and unseen failures} First, a teleoperator is $\policyTask$, controlling the end-effector pose and gripper while assisted by \ours. As shown in Fig.~\ref{fig:jenga_qualitative}, the teleoperator can freely execute safe behaviors, which require careful tilting and precise block manipulation that are non-trivial to perform. 
When erratic or OOD actions are attempted, posing a risk of tower collapse, \ours reliably intervenes to correct the behavior and maintain stability within the in-distribution region. In contrast, \baselineLatent fails to preemptively detect such boundaries due to optimistic OOD imagination, ultimately allowing high-uncertainty actions. 
Next, we quantitatively evaluate filtering by replaying $50$ failure trajectories as $\policyTask$ that result in tower collapse. The corresponding action sequences are replayed as a task policy with either \ours or \baselineLatent as the safety filter. Fig.~\ref{fig:jenga:quantitative} shows that \ours leads to lower failure rates and maintains low model uncertainty.

\section{Conclusion}

\vspace{-0.1in}
In this work, we propose \oursname, a framework for reliable latent-space safe control that \textit{unifies} reachability analysis in a latent world model with OOD detection of the world model predictions. To detect unreliable out-of-distribution imaginations of the world model, we introduce a principled method to quantify the world model’s epistemic uncertainty and calibrate a threshold. We then augment the latent space with epistemic uncertainty and perform an uncertainty-aware latent reachability analysis to synthesize a safety filter that reliably safeguards arbitrary policies from both known failures and unseen safety hazards. We demonstrate that our approach reliably identifies OOD imaginations and synthesizes an uncertainty-aware latent safety filter from an offline dataset with limited coverage, enabling safe control in complex vision-based tasks by preemptively detecting safety risks and proposing safe, in-distribution backup actions.

\clearpage
\section*{Limitations} 

\para{Component vs. System-level Safety Assurances} While our uncertainty-aware safety filter empirically can prevent both seen and unseen failures by incorporating OOD failures, it does not formally guarantee zero failure rates. 
In this work, we only provide a \textit{component-level} statistical assurance on detecting OOD transitions within the world model via conformal prediction. 
Future work should study \textit{system-level} assurances on the overall safety filter that is also influenced by our reinforcement-learning approximations in high-dimensional learned latent spaces. Moreover, our framework assumes that the system starts from an in-distribution safe initial state and that no unknown disturbances or visual distractions appear during operation. 
Therefore, for robust deployment, a system-level failure monitoring mechanism is necessary, which can reliably detect when the system loses its confidence. 
While our supplementary experiments indicate that our uncertainty measure can be leveraged for such system-level failure detection (see Sec.~\ref{sec:failure-detection}), further exploration on system-level failure detection and mitigation remains as an important future work~\cite{sinha2022system, AgiaSinhaEtAl2024}.

\para{Limited Generalizability and Reliability} Our latent safety filter relies on the capabilities of the learned world model. While recent generative world models have demonstrated promising results~\cite{zhou2024dino, agarwal2025cosmos}, the world model’s predictions can be imprecise even within in-distribution regions or fail to generalize to unseen scenarios. Although our safety filter adopts a minimally conservative approach to uncertain scenarios, its performance can be further improved with additional data. 
Future work should explore safe exploration strategies or active learning methods, using quantified epistemic uncertainty as intrinsic rewards to enhance world model generalization.

\para{Challenges in Uncertainty Quantification} While our method adopts epistemic uncertainty quantification as a proxy for detecting unreliable world model imaginations, there are several limitations to this approach. Even within regions that are nominally in-distribution, world model predictions can still be imprecise or biased, particularly in complex or stochastic systems. In other words, while a transition may be classified as in-distribution, this does not guarantee the correctness of the model’s prediction, potentially leading to an imprecise safety filter. Moreover, our uncertainty quantification assumes a Gaussian distribution over the next latent prediction, which may not hold in systems with complex, multimodal dynamics. It also adopts an ensemble as a separate module from the world model, which may not faithfully capture the model’s true uncertainty (see Sec~\ref{sec:ensembleRSSM} for further discussions). Exploring methods for faithfully detecting OOD scenarios under complex, multimodal data distributions presents an important direction for future work. Additionally, our framework and the safety Bellman equation~\eqref{eqn:discounted-safety-bellman_uncertainty} does not account for aleatoric uncertainty, and thus optimizes for the expected safety violation. Extending the framework to explicitly model aleatoric uncertainty in the latent dynamics could improve robustness, enabling latent-space safe control that better anticipates worst-case outcomes under the world model’s predictions~\cite{chapman2021risk, yu2023safe, ganai2023iterative, rigter2022ramborl, nguyen2024gameplay}.


\acknowledgments{We are grateful to Michelle Zhao for her invaluable feedback on conformal prediction. We also thank Yilin Wu and Pranay Gupta for their insightful discussions and assistance with hardware setup. This material is supported in part by the NSF CAREER Award No. 2441014. Any opinions, findings, and conclusions or recommendations expressed in this material are those of the author(s) and do not necessarily reflect the views of the National Science Foundation.}


\bibliography{mybib}

\clearpage
\appendix

\let\addcontentsline\oldaddcontentsline
\renewcommand{\contentsname}{\Large{Appendix}}
\setcounter{tocdepth}{4}
\tableofcontents
\clearpage

\section{Model Training}

In this section, we present a brief overview of the latent world model used in our work. For a more comprehensive understanding, we refer readers to the original papers~\cite{hafner2019learning, Hafner2020Dream, hafner2021mastering, hafner2023dreamerv3}.

\subsection{Latent World Model.}

\paragraph{Model Architecture} We adopt the Dreamer~\cite{hafner2023dreamerv3} framework as our latent world model. Given sequences \(\{\obs_t, \action_t, \latentfailuremargin_t\}_{t=1}^{T}\), which consist of sensor observations \(\obs_t\), action vectors \(\action_t\), and scalar failure margins \(\latentfailuremargin_t \in \{-1, 1\}\), the model defines a generative process over observations and failure margins via a latent sequence \(\{\latent_t\}_{t=1}^{T}\), under the partially observable Markov decision process (POMDP) formulation. Following the Dreamer architecture, the transition model is implemented as a Recurrent State-Space Model (RSSM)~\cite{hafner2019learning}, which predicts future latent states using either Gaussian or categorical distributions parameterized by feed-forward neural networks. The transition model outputs the prior latent state \(\hat{\latent}_t\) conditioned on the previous latent and action. The encoder then combines \(\hat{\latent}_t\) and the current observation \(\obs_t\) to produce the posterior latent \(\latent_t\):
\begin{equation*}
    \text{Encoder: } \latent_t \sim \encoder(\latent_t \mid \hat{\latent}_t, \obs_t) \quad
    \text{Transition: } \hat{\latent}_t \sim \dynz(\hat{\latent}_t \mid \latent_{t-1}, \action_{t-1}) \quad
    \text{Failure: } \latentfailuremargin_t \sim \ellz(\latentfailuremargin_t \mid \latent_t).
\end{equation*} Note that during imagination rollouts, only the prior latents are used, as observations are unavailable. The RSSM uses a Gated Recurrent Unit (GRU) to compute deterministic recurrent features, which are concatenated with samples from the stochastic state to form the full latent \(\latent_t\). Observations are decoded from the latent state using either a deconvolutional network or a multilayer perceptron (MLP), and modeled with a Gaussian likelihood. The failure classifier is trained to predict a Bernoulli likelihood over failure margins.

\paragraph{Failure Margin Function} We further assume that this offline dataset is annotated with failure labels at each timestep, indicating whether the ground-truth, but unknown, state has violated safety. These labels are used to train a margin function $\ellz( \latent_t)$ that implicitly defines a failure set in the latent space $\failure = \{ \latent \mid \ellz(\latent) < 0 \}$.

\paragraph{Loss Function} Due to the model's nonlinearity, the true posterior over latent states required for learning cannot be computed analytically. Instead, RSSM adopts a mean-field approximation that extends the framework to partially observable Markov decision processes (POMDPs). Specifically, it factorizes the variational distribution \( q(\latent_{1:T} \mid \obs_{1:T}, \action_{1:T}) \) as a product of encoder and latent dynamics terms:
\[
q(\latent_{1:T} \mid \obs_{1:T}, \action_{1:T}) = \prod_{t=1}^{T} q(\latent_t \mid \latent_{t-1}, \obs_t, \action_t),
\]
which infers an approximate posterior using past observations and actions. A variational lower bound on the data log-likelihood can then be derived using Jensen's inequality:
\begin{align}
\ln p(o_{1:T}, l_{1:T} \mid a_{1:T}) 
&\triangleq \ln \mathbb{E}_{p(\latent_{1:T} \mid a_{1:T})} \left[ \prod_{t=1}^T p(o_t, l_t \mid \latent_t) \right] \\
&\geq \mathbb{E}_{q(\latent_{1:T} \mid o_{1:T}, a_{1:T})} \left[ \sum_{t=1}^T \ln p(o_t \mid \latent_t) + \ln p(l_t \mid \latent_t) 
+ \ln p(\latent_t \mid \latent_{t-1}, a_{t-1}) - \ln q(\latent_t \mid o_{\leq t}, a_{<t}) \right] \\
&= \mathbb{E}_{q} \left[ 
\sum_{t=1}^T  
\underbrace{\ln p(o_t \mid \latent_t)}_{\text{reconstruction loss}} + \underbrace{\ln p(l_t \mid \latent_t)}_{\text{failure margin loss}} 
- \underbrace{\mathrm{KL} \left[ q(\latent_t \mid o_{\leq t}, a_{<t}) \,\|\, p(\latent_t \mid \latent_{t-1}, a_{t-1}) \right]}_{\text{KL loss}} 
\right].
\end{align} All components of the world model are optimized jointly. The encoder and failure margin function are trained to maximize the log-likelihood of their respective targets, while the dynamics model is optimized to produce latent states that facilitate these prediction tasks.

\paragraph{Implementation Details} We build on the open-source implementation of DreamerV3\footnote{\url{https://github.com/NM512/dreamerv3-torch}}. For the Dubin's Car experiments, we use a continuous stochastic latent space modeled as a 32-dimensional Gaussian. In contrast, for high-dimensional visual manipulation tasks, we adopt a discrete latent representation composed of 32 categorical variables, each with 32 classes, resulting in a 1024-dimensional stochastic latent space. Actions are represented using delta pose control—a 6-dimensional vector corresponding to normalized changes in the end-effector pose—along with an additional dimension for the gripper action. The relevant hyperparameters for Dubin's Car and visual manipulation experiments are listed in Table~\ref{tab:dubins_hyperparams} and Table~\ref{tab:manipulation_hyperparams}, respectively.
\begin{table}[ht]
\footnotesize{
    \centering
    \begin{minipage}{0.48\linewidth}
    \centering
    \renewcommand{\arraystretch}{1.2}{
    \resizebox{0.95\textwidth}{!}{
    \begin{tabular}{cc}
        \toprule
        \textbf{\textsc{Hyperparameter}} & \textbf{\textsc{Value}} \\
        \midrule
        \textsc{Image Dimension} & [128, 128, 3] \\
        \textsc{Action Dimension} & 3 (Discrete) \\
        \textsc{Stochastic Latent} & Gaussian \\
        \textsc{Latent Dim (Deterministic)} & 512 \\
        \textsc{Latent Dim (Stochastic)} & 32 \\
        \textsc{Activation Function} & SiLU \\
        \textsc{Encoder CNN Depth} & 32 \\
        \textsc{Encoder MLP Layers} & 5 \\
        \textsc{Failure Classifier Layers} & 2 \\
        \textsc{Batch Size} & 16 \\
        \textsc{Batch Length} & 32 \\
        \textsc{Optimizer} & Adam \\
        \textsc{Learning Rate} & 1e-4 \\
        \textsc{Iterations} & 100000 \\
        \bottomrule
        \end{tabular}
        }
    }
    \vspace{0.1in}
    \caption{Dubin's Car Hyperparameters}
    \label{tab:dubins_hyperparams}
    \end{minipage}
    \hfill
    \begin{minipage}{0.48\linewidth}
    \centering
    \renewcommand{\arraystretch}{1.2}{
    \resizebox{0.95\textwidth}{!}{
        \begin{tabular}{cc}
        \toprule
        \textbf{\textsc{Hyperparameter}} & \textbf{\textsc{Value}} \\
        \midrule
        \textsc{Image Dimension (Simulation)} & [2, 128, 128, 3] \\
        \textsc{Image Dimension (Real-world)} & [2, 256, 256, 3] \\
        \textsc{Proprioception Dimension} & 7 \\
        \textsc{Action Dimension} & 7 (Continuous) \\
        \textsc{Stochastic Latent} & Categorical \\
        \textsc{Latent Dim (Deterministic)} & 512 \\
        \textsc{Latent Dim (Stochastic)} & 32 $\times$ 32 \\
        \textsc{Activation Function} & SiLU \\
        \textsc{Encoder CNN Depth} & 32 \\
        \textsc{Encoder MLP Layers} & 5 \\
        \textsc{Failure Classifier Layers} & 2 \\
        \textsc{Batch Size} & 16 \\
        \textsc{Batch Length} & 64 \\
        \textsc{Optimizer} & Adam \\
        \textsc{Learning Rate} & 1e-4 \\
        \textsc{Iterations} & 200000 \\
        \bottomrule
        \end{tabular}
        }
    }
    \vspace{0.1in}
    \caption{Visual Manipulation Hyperparameters}
    \label{tab:manipulation_hyperparams}
    \end{minipage}}
\end{table}

\subsection{Probabilistic Ensemble Latent Predictor.}\label{sec:appendix_ensemble}

\begin{table}[ht]
\centering
\small
\renewcommand{\arraystretch}{1.15}{
    \begin{tabular}{lccc}
    \toprule
    \textbf{\textsc{Layer}} & \textbf{\textsc{Input Dim}} & \textbf{\textsc{Output Dim}} & \textbf{\textsc{Normalization}} \\
    \midrule
    \texttt{Linear} & \(d_\text{in}\) & \(d_\text{in}\) & \texttt{LayerNorm} \\
    \texttt{Linear} & \(d_\text{in}\) & \(2d_\text{in}\) & \texttt{LayerNorm} \\
    \texttt{Linear} & \(2d_\text{in}\) & \(3d_\text{in}\) & \texttt{LayerNorm} \\
    \texttt{Linear} & \(3d_\text{in}\) & \(d_\text{in}\) & \texttt{LayerNorm} \\
    \texttt{Linear} & \(d_\text{in}\) & \(2d_\text{out}\) & \texttt{None} \\
    \bottomrule
    \end{tabular}
}
\vspace{0.1in}
\caption{Latent Predictor Architecture}
\label{tab:penn_architecture}
\vspace{-0.2in}
\end{table}

\paragraph{Model Architecture}  
We build upon the open-source implementation provided by~\cite{kim2025learning}\footnote{\url{https://github.com/tkkim-robot/online_adaptive_cbf}}. Each ensemble member is initialized independently with random weights, resulting in diverse initializations across the ensemble. The architecture of each ensemble member in the latent predictor is summarized in Table~\ref{tab:penn_architecture}. Each member consists of five fully connected layers, initialized independently at random, and outputs a \(2d_\text{out}\)-dimensional vector corresponding to the mean and variance of a diagonal Gaussian. The input dimension \(d_\text{in}\) corresponds to the latent state, which is \(512 + 32 = 544\) for continuous Gaussian latents and \(512 + 32 \times 32 = 1536\) for discrete categorical latents. All ensemble members are trained independently and implemented using \texttt{torch.baddbmm} for efficient batch matrix multiplication during inference. The size of the ensemble is presented in Table~\ref{tab:ensemble_hyperparam}. \begin{table}[ht]
\centering
\small
\renewcommand{\arraystretch}{1.2}{
     \resizebox{0.7\textwidth}{!}{
    \begin{tabular}{cccc}
    \toprule
     & \textbf{\textsc{Dubin's Car}} & \textbf{\textsc{Block Plucking}} & \textbf{\textsc{Jenga}} \\
    \midrule
    \textsc{Ensemble Size} $\maxEnsemble$ & 10 & 5 & 5 \\
    \bottomrule
    \end{tabular}
    }
}\vspace{0.1in}
\caption{Number of ensemble members in experiments.}
\label{tab:ensemble_hyperparam}
\end{table}

\paragraph{Training} The ensemble latent predictor is trained with a frozen, pre-trained latent dynamics model, with the same dataset $\dataset_\text{train}$. Given a set of latent transitions \(\big\{\{(\latent_t, \action_t, \latent_{t+1})\}_{t=1}^{T-1}\big\}\) obtained from the offline dataset using the learned latent model, each ensemble member is trained by minimizing the Gaussian negative log-likelihood (NLL):
\begin{equation}
\mathcal{L}_{\text{train}}(\ensembleparam_k) = \sum_{t=1}^{T-1} \left[\mu_{\ensembleparam_k} - \latent_{t+1}\right]^\top \Sigma_{\ensembleparam_k}^{-1} \left[\mu_{\ensembleparam_k} - \latent_{t+1}\right] + \log \det \Sigma_{\ensembleparam_k}.
\label{eq:train_loss}
\end{equation}

\paragraph{A Brief Background on Jensen-Rényi Divergence}
To quantify epistemic uncertainty from ensemble disagreement, it is essential to distinguish it from aleatoric uncertainty. Without this separation, it becomes unclear whether the uncertainty is from a lack of model knowledge or from inherent, irreducible system stochasticity, such as model ambiguity or sensor noise. A common approach for measuring disagreement between predictive distributions is the Kullback–Leibler (KL) divergence. However, KL divergence is asymmetric and limited to pairwise comparisons, making it unsuitable for capturing ensemble-wide disagreement.

The Jensen-Rényi divergence (JRD) extends the well-known Jensen-Shannon divergence with Rényi entropy $H_{\alpha}(Z)$ of a random variable $Z$: \begin{equation}
    \operatorname{JRD}\left( \ensemble_{1:\maxEnsemble}\right) \triangleq  H_{\alpha}\left(\sum_{k=1}^{K} \frac{1}{K} \ensemble_{k} \right)- \sum_{k=1}^{\maxEnsemble} \frac{1}{K} H_{\alpha}\left(\ensemble_{k}\right), \quad  H_{\alpha}(Z)=\frac{1}{1-\alpha} \log \int p(z)^{\alpha} \mathrm{d}x.
\end{equation} 
However, computing JRD is intractable as it involves estimating the entropy of a mixture of Gaussians, which has no analytical solution. Although JRD can be estimated via Monte Carlo sampling, such approximations are computationally expensive and impractical for real-time applications. To address this, Wang et al.\cite{wang2009closed} introduced a closed-form JRD formulation based on quadratic Rényi entropy ($\alpha=2$)~\cite{wang2009closed}, enabling efficient and analytic computation of the divergence among Gaussian mixture models (GMMs):
\begin{equation}
\begin{aligned}
    \operatorname{JRD}\left( \ensemble_{1:\maxEnsemble}\right) &= - \log \Biggl[  \frac{1}{\maxEnsemble^{2}}  \sum_{i, j}^{\maxEnsemble} \mathfrak{D}\left(\ensemble_{i}, \ensemble_{j}\right)\Biggr] + \frac{1}{\maxEnsemble} \sum_{i}^{\maxEnsemble} \log \left[ \mathfrak{D}\left(\ensemble_{i}, \ensemble_{i}\right) \right], \quad  \text{where}\\
\mathfrak{D}\left(\ensemble_{i}, \ensemble_{j}\right)&=\frac{1}{|\Phi|^{\frac{1}{2}}} \exp \left(-\frac{1}{2} \Delta^{\top} \Phi^{-1} \Delta\right) \quad \text{with} \quad \Phi=\Sigma_{\phi_i}+\Sigma_{\phi_j} \quad \text{and} \quad \Delta=\mu_{\phi_i}-\mu_{\phi_j}.
\end{aligned}
\end{equation}

We refer readers to \cite{wang2009closed} for a detailed explanation of closed-form JRD, and its practical applications in learning-based settings~\cite{shyam2019model, kim2023bridging}.

\subsection{Why not ensemble the latent dynamics model itself?}\label{sec:ensembleRSSM} While prior works often ensemble the dynamics model directly to estimate epistemic uncertainty~\cite{chua2018deep, shyam2019model, kim2023bridging}, we instead introduce a separate ensemble of latent predictors as a proxy for uncertainty estimation of the learned latent world model. Although a detailed analysis of this design choice is beyond the scope of our contributions, we briefly address several challenges of ensembling the latent dynamics model and give justification for our design choice.

\paragraph{Practical Challenges of Ensembling RSSM} Since the latent world model jointly optimizes both the latent representation and the transition dynamics, ensembling this model would require optimizing over a non-stationary and noisy target—the next latent state $\latent_{t+1}$—which can lead to training instability. Furthermore, because the latent dynamics are trained via distribution matching rather than a direct regression objective, it becomes intractable to apply standard ensemble training techniques. Although recent works~\cite{rafailov2021offline, seyde2022learning, as2025actsafe} attempt to circumvent this issue by randomly sampling a single dynamics member during training, we empirically find that this approach results in weaker representation learning and less reliable uncertainty estimation compared to our method (See \ref{sec:dubins_ood_ablations}). Additionally, the stochastic latent variables in the world model are optimized for sampling within the latent space to enable next-state prediction, rather than for explicitly modeling aleatoric uncertainty. We empirically find that the predicted variance or distribution produced by RSSM does not faithfully capture the calibrated aleatoric uncertainty inherent in the dynamics, which is critical for accurate uncertainty quantification and for distinguishing aleatoric from epistemic uncertainty.

\paragraph{Separate Ensemble Module Enables Efficient Uncertainty Quantification} Lastly, our design choice enables efficient epistemic uncertainty quantification for the latent world model. As the latent world model tends to be significantly larger than typical state-based dynamics models, employing a lightweight, separate module offers a practical and scalable way to capture uncertainty. This is especially important as recent generative world models, such as~\cite{micheli2023transformers, zhou2024dino}, continue to grow in size, making it infeasible to ensemble the model itself. Note that our method is capable of forwarding both the latent representation and its associated uncertainty in under 0.1 seconds on a standard desktop setup, using approximately $2-4$GB of VRAM for the ensemble.
\section{Latent-Space Reachability Anaylsis}

\subsection{A Brief Background on HJ Reachability} Hamilton-Jacobi (HJ) reachability is a control-theoretic framework for safety analysis that identifies when current actions may lead to future failures and computes best-effort policies to mitigate such outcomes~\cite{mitchell2005time, hsu2023safety}. Given a dynamical system with state \(\state \in \stateSpace\), action \(\action \in \actionSpace\), and dynamics \(\state_{t+1} = \dyns(\state_t, \action_t)\), HJ reachability seeks to determine the safe set that can prevent the system from entering a designated \textit{failure set} \(\failure = \{ \state \; | \; \marginfunc(\state) < 0 \}\), which is represented by a margin function \(\marginfunc: \stateSpace \rightarrow \mathbb{R}\). The framework aims to find the \textit{unsafe set}, denoted \(\unsafeSet \subset \stateSpace\), which includes all states from which the system is inevitably driven into \(\failure\) despite the best effort, and the best effort safety-preserving policy to avoid entering the unsafe set.

The framework jointly computes (\romannumeral 1) a safety value function \(\monitor: \stateSpace \rightarrow \mathbb{R}\), which quantifies the minimal safety margin the system can achieve from a given state \(\state\) under optimal behavior, and  (\romannumeral 2) a best-effort safety-preserving policy \(\fallback: \stateSpace \rightarrow \actionSpace\).  
These are obtained by solving an optimal control problem governed by the following fixed-point safety Bellman equation:
\begin{equation}
\valfunc(\state) = \min \Big\{ \marginfunc(\state), ~ \max_{\action \in \actionSpace} \valfunc(\dyns(\state, \action)) \Big\}, \quad \fallback(\state) := \arg\max_{\action \in \actionSpace} \valfunc(\dyns(\state, \action)).
\label{eq:trad-fixed-pt-bellman}
\end{equation}

To tractably approximate solutions to high-dimensional reachability problems, \citet{fisac2019bridging} propose using reinforcement learning by replacing the standard Bellman equation for cumulative reward with a time-discounted counterpart of Eq.~\ref{eq:trad-fixed-pt-bellman}:
\begin{equation}
        \label{eqn:discounted-reachability-bellman}
        \valfunc(\state_t) = (1-\gamma)\marginfunc(\state) + \gamma \min \Big\{ \marginfunc_{\theta}(\state), \max_{\action \in \actionSpace}     \valfunc \big(\dynamics(\state, \action) \big) \Big\},
\end{equation} where \(\gamma\) is the discount factor that ensures contraction of the Bellman operator. The resulting \textit{unsafe set}, denoted \(\unsafeSet \subset \stateSpace\), captures all states from which the system can no longer avoid entering \(\failure\), and is defined as the zero sublevel set of the value function: \(\unsafeSet := \{ \state \; | \; \valfunc(\state) < 0 \}.\) At deployment time, the safety value function and safety policy enable \textit{safety filtering}: detecting unsafe actions proposed by any task policy \(\policyTask\) and minimally adjusting them only when necessary to ensure the system remains within the safe set. We refer readers to survey papers for further details~\cite{hsu2023safety, wabersich2023data}.

\subsection{Implementation Details.}\label{sec:appendix_implementation_details} To approximate the safety filter via reinforcement learning, we adopt training strategies from model-based reinforcement learning, enabling reachability analysis entirely through latent imagination. Using the learned dynamics model, we initialize rollouts by encoding randomly sampled data from the offline training dataset into the latent space. Starting from these initial latent states, imagined rollouts are used as a simulated environment for off-policy reinforcement learning. Note that during imagination rollouts, only the prior latents are used, as observations are unavailable

\paragraph{Discrete Action Space} For the Dubins Car experiments, which operate in a discrete action space, we use Double DQN (DDQN)~\cite{van2016deep} to train the safety value function. The Q-function is implemented as a 3-layer multilayer perceptron~(MLP) with a hidden dimension of 100, producing Q-values for each of the 3 discrete actions. The associated hyperparameters are summarized in Table.~\ref{tab:DDQN_hyperparams}.

\begin{table}[h]
    \footnotesize
     \begin{minipage}{0.45\linewidth}
    \renewcommand{\arraystretch}{1.3}
    \resizebox{0.95\textwidth}{!}{
    \begin{tabular}{lc}
        \toprule
        \textsc{\textbf{Hyperparameter}} & \textsc{\textbf{Value}} \\
        \midrule
        \textsc{Architecture} & [100, 100] \\
        \textsc{Learning Rate} & 1e-3 \\
        \textsc{Optimizer} & AdamW \\
        \textsc{Discount Factor} $\gamma$ & 0.9999 \\
        \textsc{Num Iterations} & 50000 \\
        \textsc{Memory Buffer Size} & 20000 \\
        \textsc{Batch Size} & 256 \\
        \textsc{Max Imagination Steps} & 20 \\
        \bottomrule
    \end{tabular}
    }
    \vspace{0.1in}
    \caption{DDQN Hyperparameters}
    \label{tab:DDQN_hyperparams}
    \end{minipage}
    \hfill
    \begin{minipage}{0.5\linewidth} 
    \resizebox{0.95\textwidth}{!}{
    \renewcommand{\arraystretch}{1.2}
    \begin{tabular}{lc}
        \toprule
        \textsc{\textbf{Hyperparameter}} & \textsc{\textbf{Value}} \\
        \midrule
        \textsc{Actor Architecture} & [512, 512, 512, 512] \\
        \textsc{Critic Architecture} & [512, 512, 512, 512] \\
        \textsc{Normalization} & LayerNorm \\
        \textsc{Activation} & ReLU \\
        \textsc{Discount Factor} $\gamma$ & 0.85 $\rightarrow$ 0.9999 \\
        \textsc{Learning Rate (Critic)} & 1e-4 \\
        \textsc{Learning Rate (Actor)} & 1e-4 \\
        \textsc{Optimizer} & AdamW \\
        \textsc{Number of Iterations} & 200000 \\
        \textsc{Replay Buffer Size} & 500000 \\
        \textsc{Batch Size} & 512 \\
        \textsc{Max Imagination Steps} & 30 \\
        \bottomrule
    \end{tabular}
    }
    \vspace{0.1in}
    \caption{SAC hyperparameters.}
    \label{tab:SAC_hyperparams}
    \end{minipage}
\end{table}

Importantly, in the Dubins car task, the vehicle is tasked with avoiding a target located at the center, and its trajectory may extend beyond the bounding box, which is safe, while regions outside the box are highly uncertain and out-of-distribution. To avoid overly conservative behavior that penalizes successful avoidance, we track the true state of the vehicle and omit the OOD penalty for states outside the bounding box, defined as those with \(|p_x| > 1\) or \(|p_y| > 1\).

\paragraph{Continuous Action Space} For continuous control in both simulated and real-world visual manipulation tasks, we adopt Soft Actor-Critic (SAC)~\cite{haarnoja2018soft} within an off-policy, model-based reinforcement learning framework. We model the safety value function as a latent-action value function \( Q(\latent, \action)\) that is conditioned on the action. The safety policy is parameterized by an actor-network \(\action \sim \fallback(\cdot \mid \latent)\). The safety value can be evaluated by $\monitor(\latent) = \max_aQ(\latent, \action) = Q(\latent,\fallback(\latent))$.

At each time step, we store imagined transitions \((\latentUncertainty, \action, \latentfailuremargin, \latentUncertainty')\) in the replay buffer $\mathcal{B}$, where \(\latentfailuremargin\) is given by the safety margin function \(\marginfuncUncertainty(\latentUncertainty, \uncertainty)\) as defined in Eq.~\ref{eq:uncertainty-aware-latent}. We then optimize the critic using the following objectives:
\begin{equation}
       \mathcal{L}_\text{critic} := \mathbb{E}_{(\latentUncertainty, \action, r, \latentUncertainty',) \sim \mathcal{B}} \left[ \left(Q(\latent, \action) - y\right)^2 \right], \quad 
y = (1 - \gamma)\, r + \gamma \min\{r, \max_{\action'} Q(\latent', \action')\}. 
\end{equation} Note that we do not parametrize the uncertainty variable $\uncertainty$ directly in the value function but intrinsically leverage it with the safety margin function. To stabilize training, we maintain two Q-functions and use target networks for the temporal difference updates. The policy is optimized following the policy gradient induced by the critic and entropy loss term:
\begin{equation}
    \mathcal{L}_{\text{actor}} := \mathbb{E}_{\latent \sim \mathcal{B}} \left[ -Q(\latent, {\action}) + \beta \log \fallback({\action} \mid \latent) \right], \quad {\action} \sim \fallback(\cdot \mid \latent),
\end{equation} where $\gamma$ is scheduled from $0.85$ to $0.9999$ and $\beta$ is the hyperparameter for exploration. These updates enable the actor to select actions that maximize expected safety margins while the critic estimates the corresponding safety values under uncertainty. The hyperparameters for training SAC are summarized in Table.~\ref{tab:SAC_hyperparams}.

\subsection{Uncertainty-aware Safety Filter.} To implement the safety filter in the uncertainty-aware latent space described in Eq.~\ref{eq:uncertainty-aware-latent}, we leverage the trained action-conditioned safety value function \(Q\). At each time step, we evaluate whether executing the action proposed by the task policy \(\policyTask\) leads to an unsafe outcome, as determined by reachability analysis, and, if so, override it using the safety policy \(\fallback\).  

In practice, the filtering condition \( \valfuncUncertainty(\latentUncertainty') \le \delta \), which evaluates on the predicted next latent $\latentUncertainty' \sim \dynamicsUncertainty (\latentUncertainty, \policyTask)$, is assessed by monitoring two criteria: (\romannumeral 1) if the epistemic uncertainty exceeds the threshold (i.e., \(\disagreement(\latent, \policyTask) > \threshold\)); or (\romannumeral 2) the safety value is low (i.e. \(Q(z', \fallback(z')) \le \delta\)). The union of these conditions indicates that the safety value of the next state is below the safe margin. Recall the definition of the uncertainty-aware safety margin function~(\ref{eq:uncertainty-aware-latent}) and the Bellman update~(\ref{eqn:discounted-safety-bellman_uncertainty}):
\begin{equation*}
   \marginfuncUncertainty(\latentUncertainty_t) = \min\big\{ \ellz(\latent_t)\,,  \OODpenalty \, (\threshold - \uncertainty_t)\big\}, \quad \valfuncUncertainty(\latent, \uncertainty) = \min\left\{ \marginfuncUncertainty(\latent, \uncertainty), \max_{\action} \valfuncUncertainty(\latent', \uncertainty') \right\},
\end{equation*} where $\latent', \uncertainty'$ is sampled from $\dynamicsUncertainty(\latent, \action)$. When the disagreement is above the threshold, the uncertainty triggers the OOD penalty, which yields \(\marginfuncUncertainty(\latentUncertainty) = -\OODpenalty < 0 \), ensuring that the value function is below the safety filter margin $ \delta > 0$. Alternatively, if the value at the next latent state is small, i.e., \(Q(z', \pi(z')) \le \delta\), then the second term in the Bellman update becomes small, also driving \(\valfuncUncertainty\left( \latentUncertainty' \right)  \le \delta\). Hence, either condition implies that the next transition is unsafe under the safety filter:
\[
\disagreement(\latent, \policyTask) > \threshold \quad \text{or} \quad Q(z', \pi(z')) \le \delta \quad \Longleftrightarrow \quad \valfuncUncertainty\left( \latentUncertainty' \right) \le \delta.
\]

\subsection{Failure Detection of the Uncertainty-aware Safety Filter}\label{sec:failure-detection}
\paragraph{Does our safety filter always guarantee safety?} Our framework assumes that the robot starts from an in-distribution initial state and maintains approximate control invariance with respect to an estimated safe set in the latent space. However, since the safety filter is trained via reinforcement learning and relies on an imperfect latent dynamics model, safety cannot be guaranteed in all cases. The reliability of the learned filter can degrade in several situations—for instance, when the system begins in an out-of-distribution state (e.g., due to an OOD visual input at test time) or when the filter fails to prevent transitions into unsafe regions. In such cases, the safety filter may behave unpredictably, executing random or overconfident actions or even exacerbating unsafe situations. To ensure safe deployment, it is essential to detect when the safety filter becomes unreliable. In such cases, the system should halt and request human intervention. Without this safeguard, the robot may continue operating despite its internal safety mechanism failing. 

\paragraph{System-level Failure Detection} Failures of learned safety filters can arise from a range of sources, including OOD sensory inputs, misspecified dynamics models, or inaccurately learned safety value functions. A reliable safety filter should exhibit consistent behavior under bounded epistemic uncertainty. To detect violations of this principle, we monitor whether the backup action \(\fallback(\latent)\) leads to a transition with sufficiently low predictive uncertainty. If it does not, we assume the system has entered the OOD failure set and must stop operation.

Based on the safety filtering rule in Eq.~\ref{eq:latent-safe-control}, the selected action is expected to avoid transitions that induce high predictive uncertainty. Formally, the safety filter should satisfy:
\(\disagreement\left(\latent, \action^\text{exec} )\right) \le \threshold.\) Conversely, if the filtered action itself leads to excessive epistemic uncertainty, we consider the system to have entered the unsafe set, which the robot cannot automatically recover from. In this case, the safety guarantees provided by the filter no longer hold, and the system should halt operation. In particular, if even the fallback action \(\fallback(\latent)\) results in high disagreement, the system is deemed unrecoverable under the current safety filter: \(\disagreement(\latent, \fallback(\latent)) > \threshold.\) This motivates a modification to the filtering rule, introducing an explicit halting condition when the filter is unable to guarantee a safe and confident action. With predicted next latent state $\latentUncertainty' \sim \dynamicsUncertainty(\latentUncertainty, \policyTask)$ the filter is constructed as:
\begin{equation}\label{eq:modified-uncertainty-aware-safety-filter}
\phi\left(\latentUncertainty, \policyTask\right) := 
\begin{cases}
\policyTask, & \text{if } \valfuncUncertainty\left(\latentUncertainty')\right) > \delta, \\ \textcolor{green}{\fallback(\latentUncertainty)}, & \text{if } \valfuncUncertainty\left(\latentUncertainty')\right) \le \delta \,\, \text{and} \,\, \disagreement(\latent, \fallback(\latent)) \le \threshold, \\
\textcolor{red}{\text{HALT}}, & \text{otherwise.}
\end{cases}
\end{equation}

\begin{figure}[ht]
    \centering
    \includegraphics[width=1.0\linewidth]{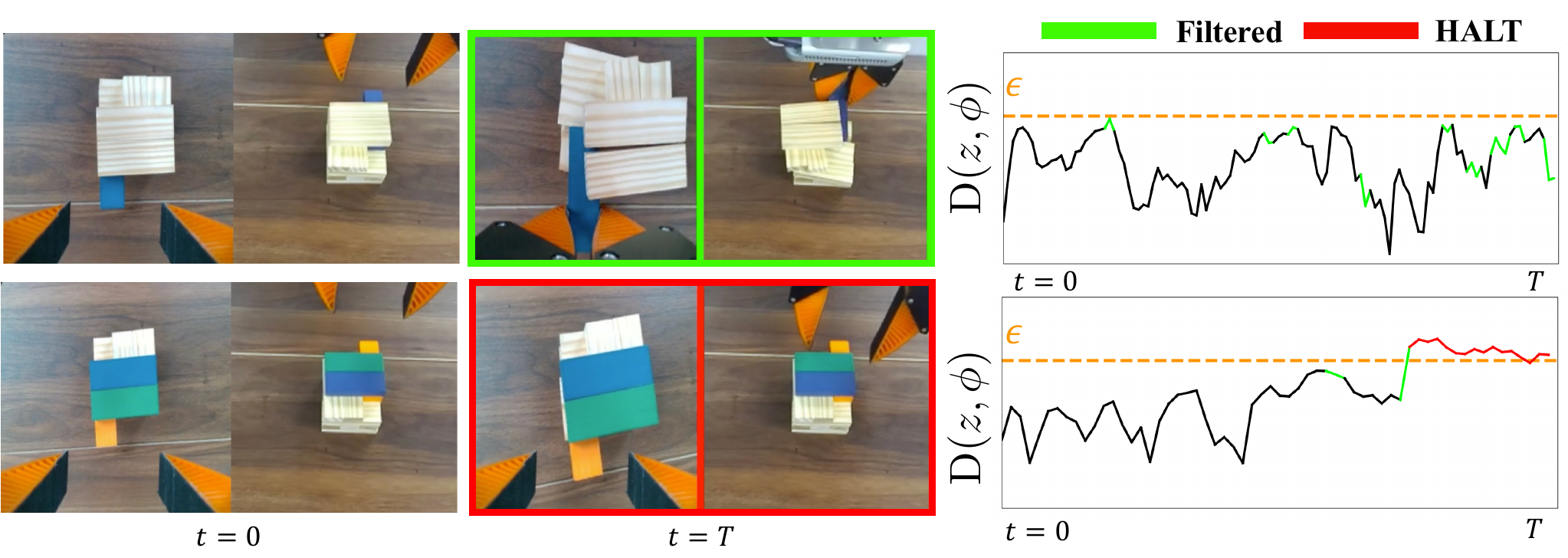}
    \caption{\textit{Top row}: despite a color change in the target block, the latent dynamics model remains reliable, maintaining predictive uncertainty below the threshold. \textit{Bottom row}: in contrast, when the visual input deviates significantly from the training distribution, the model becomes unreliable. The safety filter fails to maintain predictive uncertainty below the threshold, prompting the system to halt in order to avoid actions that could compromise or aggravate safety.}
    \label{fig:halt_ablation}
\end{figure}

\paragraph{Results: OOD Visual inputs} Fig.~\ref{fig:halt_ablation} illustrates the outcome of failure detection by the safety filter in the Jenga task. In this scenario, a teleoperator attempts to grasp a block and executes an unsafe action—pushing the block to the right. The learned safety filter intervenes to suppress this unsafe behavior. Although the block colors differ from those encountered during training, such visual changes do not inherently indicate out-of-distribution inputs. Instead, the decision to halt is governed by the reliability of the filtering system. When the color of the target block changes but remains within the model’s generalization capacity, the latent dynamics model remains reliable, maintaining predictive uncertainty below the threshold. In contrast, when the visual input deviates substantially from the training distribution, the model becomes unreliable. The safety filter then fails to keep uncertainty within acceptable bounds, prompting the system to halt in order to prevent potentially dangerous actions.

\begin{figure}[h]
    \centering
    \includegraphics[width=1.0\linewidth]{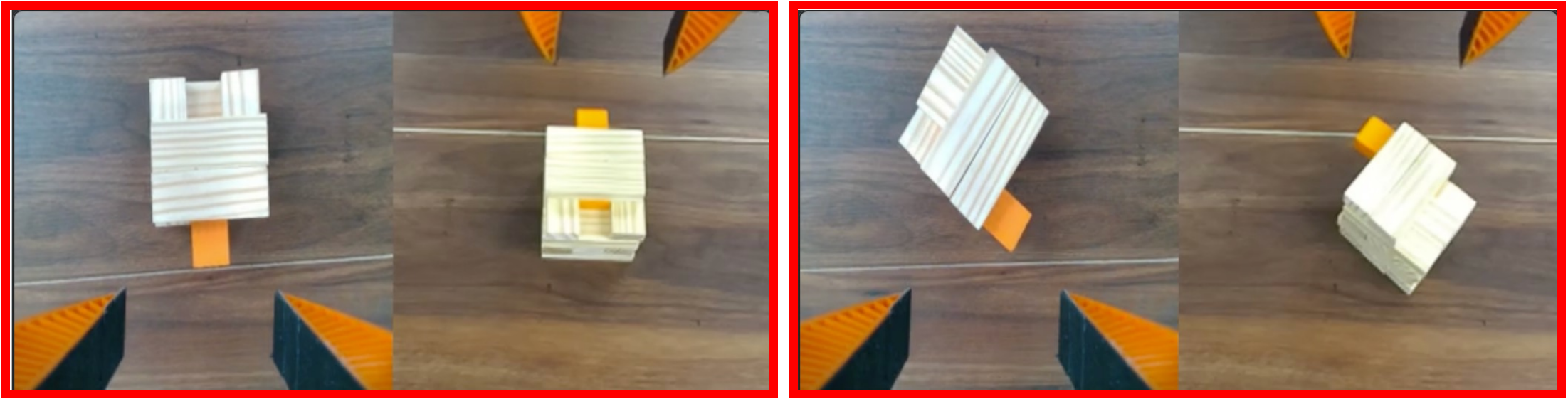}
    \caption{OOD settings that lead the system to halt.}
    \label{fig:halt_ood}
\end{figure} Fig.~\ref{fig:halt_ood} shows additional scenarios where the system safely halts upon detecting unrecoverable conditions due to OOD inputs that differ significantly from the training data.

\subsection{Brief Backgrounds on Offline Reinforcement Learning} Offline reinforcement learning (RL) learns policies from a static dataset of past interactions, making it well-suited for applications where online exploration poses safety risks~\cite{levine2020offline, recoveryrl, urpi2021risk, kolev2024efficient}. A major challenge in offline RL is the distribution shift between the learned policy and the behavior policy that collected the data~\cite{jin2021pessimism, sun2023model}, which often leads to overestimation of policy evaluations on OOD scenarios~\cite{bai2022pessimistic}. To address this, conservatism is introduced by penalizing value functions, preventing over-optimism on OOD actions~\cite{kumar2020conservative, yu2021combo, rigter2022ramborl}. In offline model-based RL (MBRL), a dynamics model is learned from the static dataset and used to generate synthetic data for policy learning~\cite{chua2018deep, buckman2018sample, yu2020mopo, kidambi2020morel, yu2021combo, sun2023model}. By quantifying the uncertainty of the learned dynamics model, these methods mitigate model exploitation and discourage the system from entering OOD scenarios. Inspired by this, our work quantifies uncertainty in a learned latent dynamics model and ensures a safety filter to proactively prevent the system from entering OOD regions.
\section{Conformal Prediction for Calibrating OOD Threshold}\label{sec:appendix_conformal}

In this section, we briefly introduce conformal prediction and outline the details of our calibration procedure. For a comprehensive overview, we refer readers to~\cite{vovk2005algorithmic, angelopoulos2023conformal}. Conformal prediction is a statistically principled framework for constructing prediction sets or regions with guaranteed coverage, relying only on mild assumptions such as data exchangeability or i.i.d. sampling. Given a user-specified significance level \(\alpha\), it guarantees that the true target lies within the constructed prediction region with probability at least \(1 - \alpha\).

Let \(\{P_1, P_2, \dots, P_N\}\) be a set of \(N\) i.i.d. nonconformity scores. The goal is to compute a threshold \(C\) such that a new test sample is included in the prediction region with high probability. Conformal prediction provides the following guarantee:
\[
\mathbb{P}(P_{\text{test}} \leq C) \geq 1 - \alpha.
\]
The threshold \(C\) is typically chosen as the empirical \((1 - \alpha)\)-quantile of the calibration scores. This is computed by sorting the set \(\{P_1, \dots, P_N\}\) and selecting the \(\lceil (1 - \alpha)(N + 1) \rceil\)-th smallest value, where \(\lceil \cdot \rceil\) denotes the ceiling function.

\subsection{Trajectory-level Conformal Prediction} Eq.~\ref{eq:coformal_guarantee} provides a probabilistic guarantee on the recall of in-distribution transitions—or, equivalently, a bound on the false positive rate. Rewriting the equation:
\begin{equation}\label{eq:quantile_guarantee}
\mathbb{P}\Big( Q_{\trajectory_{\text{test}}}^{\alphaQuantile} \le \threshold \mid \trajectory_{\text{test}} \in \dataset_\indist \Big) \ge 1 - \alphaCalibration,
\end{equation} this implies that the quantile of ensemble disagreement within a test trajectory is less than or equal to the threshold \(\threshold\) with probability at least \(1 - \alphaCalibration\). In other words, for an in-distribution test trajectory, at least a fraction \(1 - \alphaQuantile\) of its transitions are expected to have disagreement scores below \(\threshold\) with high confidence. Note that the maximum is a special case of the quantile function with \(\alphaQuantile = 0\): 
\begin{equation}
\mathbb{P}\Big( \max_t \disagreement(\latent_t^\text{test}, \action_t^\text{test}) \le \threshold \mid (\latent_t^\text{test}, \action_t^\text{test}) \in \dataset_\indist \Big) \ge 1 - \alphaCalibration,
\end{equation} where all transitions within a trajectory must fall below the threshold. Similar to~\cite{ren2023robots}, this trajectory-level in-distribution prediction set \(\mathcal{C}(\trajectory_\text{test})=\Big\{ \trajectory_{i}: Q_{\trajectory_i}^{\alphaQuantile}\leq\threshold\Big\},\) enables causal reconstruction of a transition-level in-distribution set \(\mathcal{C}(\latent_\text{test}, \action_\text{test})=\Big\{ (\latent_t, \action_t): \disagreement(\latent_t, \action_t)\leq\threshold\Big\},\) since:
\begin{equation}
\max_t \disagreement(\latent_t^\text{test}, \action_t^\text{test}) \le \threshold \ \Longleftrightarrow \ \disagreement(\latent_t^\text{test}, \action_t^\text{test}) \le \threshold \quad \forall t \in [T].
\end{equation}

However, this strict formulation tends to produce overly optimistic thresholds in practice, resulting in a high false negative rate—that is, misclassifying OOD transitions as in-distribution. This is largely due to noise and imperfection in the ensemble-based disagreement estimates. To mitigate this, we adopt a quantile-based nonconformity score, which allows for a small, controlled level of transition-level misclassification \((1 - \alphaQuantile)\) within each trajectory.

\subsection{Dataset-conditional Guarantee.}
Equations \ref{eq:coformal_guarantee} and \ref{eq:quantile_guarantee} hold marginally, with the probability taken over both the sampling of the test data and the calibration data~\cite{chakraborty2024enhancing}. However, by fixing the calibration dataset, which is drawn i.i.d.\ from \(\dataset_\indist\), we obtain a dataset-conditional guarantee~\cite{shafer2008tutorial}. Specifically, conditioned on a calibration dataset \(\dataset_\text{cal} \subset \dataset_\indist\), the coverage achieved by conformal prediction follows a Beta distribution~\cite{vovk2012conditional}:
\begin{equation}
    \mathbb{P}\Big( Q_{\tau_{\text{test}}}^{\alphaQuantile} \le \threshold  \mid \trajectory_{\text{test}} \in \dataset_\indist\Big) \sim \text{Beta}(N+1-C, C), \quad \text{where} \quad C := \lfloor (N+1) \alphaQuantile \rfloor.
\end{equation}

\subsection{Implementation Details.}

For each task, we collect a calibration dataset to determine the OOD threshold based on ensemble disagreement. This calibration dataset is a held-out subset collected alongside the training data, but it is not used during model training. Table~\ref{tab:conformal_params} summarizes the calibration dataset sizes and the conformal prediction hyperparameters used for each task.
\begin{table}[h]
\footnotesize
\centering
\resizebox{0.6\textwidth}{!}{
\renewcommand{\arraystretch}{1.2}{
    \begin{tabular}{cccc}
    \toprule
    \textsc{Task} & \textsc{Calibration Set Size} (\(N\)) & \(\alphaCalibration\) & \(\alphaQuantile\) \\
    \midrule
    \textsc{Dubin's Car} & 500 & 0.05 & 0.05 \\
    \textsc{Block Plucking} & 100 & 0.05 & 0.05 \\
    \textsc{Jenga} & 30 & 0.10 & 0.10 \\
    \bottomrule
    \end{tabular}
}}
\vspace{0.05in}
\caption{Conformal Prediction Parameters for Each Task}
\label{tab:conformal_params}
\end{table}

\section{Experiment Details.}
\subsection{Dubins Car}\label{sec:appendix_experiments_details}
\begin{figure}[h]
\centering
    \includegraphics[width=0.9\linewidth]{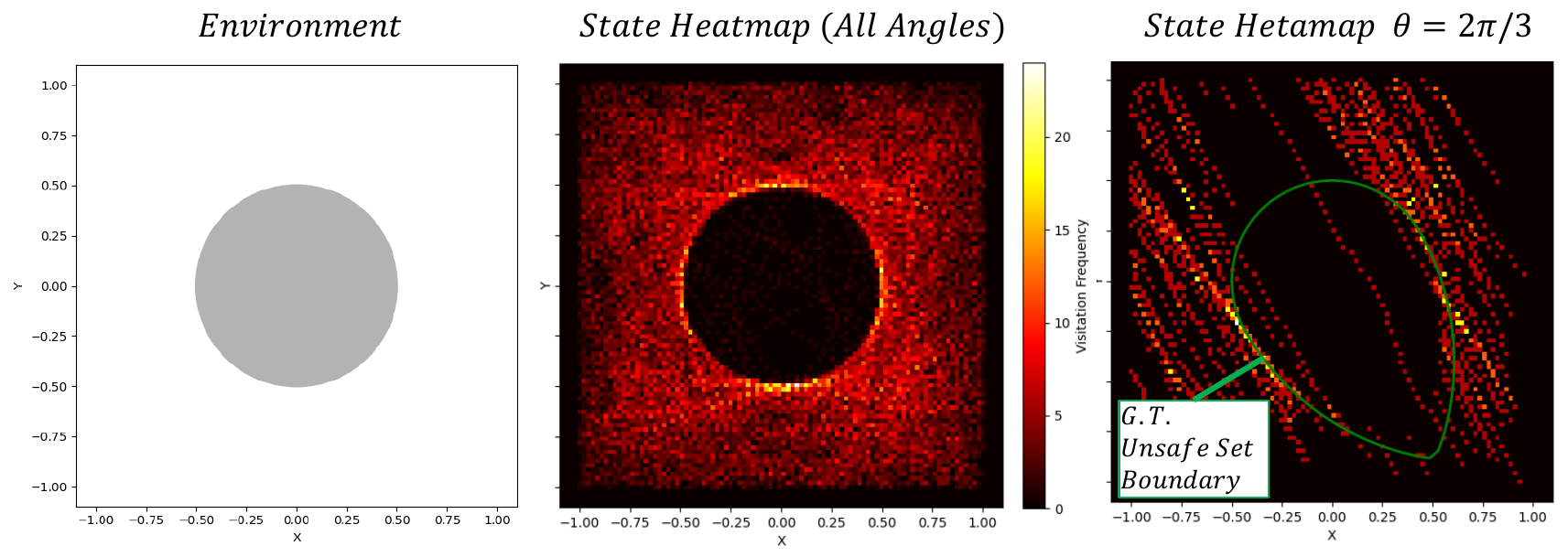}
    \caption{Visualization of the dataset consisting of expert trajectories and a few random trajectories. \textit{Left}: Environment. The gray circle at the center denotes the failure set, which expert trajectories consistently avoid. \textit{Middle}: State visitation heatmap based on (x, y) positions, showing that most data is concentrated in safe regions. \textit{Right}: Heatmap of states with a specific heading angle. Most of the data lies outside the ground-truth unsafe set, resulting in high model uncertainty around the true failure region.}
    \label{fig:expert_heatmap}
    \vspace{-0.05in}
\end{figure}

\paragraph{Expert Trajectories} The expert trajectories never enter the failure set. Importantly, this does not mean they simply stop at the boundary; rather, the vehicle also avoids entering the ground-truth unsafe set—states from which failure is inevitable despite being currently safe. These trajectories are generated using the ground-truth safety value function, computed via a grid-based solver~\cite{mitchell2007toolbox}. At the boundary of the unsafe set, the vehicle executes only safe actions, whereas outside this region, it performs random actions. As shown in Fig.~\ref{fig:expert_heatmap}, the dataset exhibits low density within the unsafe region, resulting in high epistemic uncertainty in those areas and leading to overly optimistic world model imagination near critical decision boundaries.

\paragraph{Evaluation} Since the ground-truth dynamics are known, we can compute the exact safety value function using traditional grid-based methods~\cite{mitchell2007toolbox}. This enables us to evaluate the accuracy of the safety filter’s monitor by comparing its safe/unsafe classifications against the ground truth. We compute the value functions over all three dimensions for both safety monitoring and policy evaluation. To assess the effectiveness of the learned safety policies, we evaluate whether they can successfully steer the system away from failure. Feasible initial conditions are identified using the ground-truth state-based value function, yielding approximately 10,000 candidates. To highlight challenging scenarios, we additionally report results on a curated subset of 181 cases where the system starts in a safe region but is oriented toward the failure set. Table~\ref{tab:appendix_uq_comparison} presents a quantitative comparison between two settings: one using an uncertainty-aware latent space that incorporates OOD failures, and one using latent dynamics without explicit uncertainty modeling.

\begin{table}[h]
    \centering
    \renewcommand{\arraystretch}{1.1}
    \resizebox{0.8\textwidth}{!}{
    \begin{tabular}{lccccc}
        \toprule
        & \textbf{FPR$\downarrow$} & \textbf{Pre.$\uparrow$} & \textbf{B.Acc.$\uparrow$} & \textbf{Safe (Total) $\uparrow$} & \textbf{Safe (Challenging) $\uparrow$}  \\
        \midrule
        \baselineLatent (w.o. $\failure_\ood$)  & 0.30 & 0.92 & 0.84 &  0.98 &0.63 \\
        \ours (w. $\failure_\ood$) & 0.05 & 0.98 & 0.94 &  0.97 &0.82 \\
        \bottomrule
    \end{tabular}
    }
    \vspace{0.05in}
    \caption{Safety filter performance on the Dubins Car experiment with expert trajectories.}
    \label{tab:appendix_uq_comparison}
    \vspace{-0.05in}
\end{table}

\paragraph{Dubins Car without Failure Trajectories}\label{sec:dubins_ood_ablations} To further validate the reliability of uncertainty quantification, we quantitatively compare several UQ methods applicable to the latent dynamics model in the Dubins Car setting with only OOD failures (see Sec.~\ref{sec:dubins}). Using the same offline dataset, we train latent dynamics models with different UQ methods and synthesize corresponding safety filters. For each method, the threshold is calibrated using the same held-out calibration dataset. We then evaluate the learned safety value functions against the ground-truth safety value function exactly computed with grid-based methods. Additionally, we perform closed-loop rollouts from random initial states sampled across all three dimensions of the Dubins Car state space, measuring the resulting safety rates of full trajectories. Results are summarized in Table~\ref{tab:uq_methods_comparison}.

\begin{table}[ht]
    \scriptsize
    \centering
    \resizebox{0.8\textwidth}{!}{
    \renewcommand{\arraystretch}{1.2}{
    \begin{tabular}{lcccccc}
        \toprule
        \textbf{Method} & \textbf{FPR$\downarrow$} & \textbf{Recall $\uparrow$} & \textbf{Pre.$\uparrow$} & \textbf{\boldmath$F_1\uparrow$} & \textbf{B.Acc.$\uparrow$} & \textbf{Safe Rate$\uparrow$} \\
        \midrule
        \unitVar  & 0.028 & 0.888 & 0.965 & 0.926  & 0.930 & 0.938 \\
        \textit{EnsembleRSSM} & 0.024 & 0.846 & 0.969 & 0.904  & 0.911 & 0.925 \\
        \maxVar   & 0.119 & 0.785 & 0.854 & 0.819 & 0.834 & 0.790 \\
        \density \,($\latent, \action$) & 0.123 & 0.981 & 0.876 & 0.925 & 0.929 & 0.769 \\
        \density \,($\latent$) & 0.438 & 0.993 & 0.799 & 0.667 & 0.778 & 0.640 \\
        \midrule
        \textit{JRD} ($\epsilon + 0.3$) & 0.561 & 0.987  & 0.609 & 0.753  & 0.712 & - \\
        \textit{JRD} ($\epsilon - 0.3$) & 0.033 & 0.852 & 0.957 & 0.902 & 0.909 & - \\
        \textit{JRD} & 0.049 & 0.931 &0.944 & 0.937  & 0.941 & 0.967 \\
        \bottomrule
    \end{tabular}
    }} \vspace{0.05in}
    \caption{Performance comparison of different uncertainty quantification methods.}
    \label{tab:uq_methods_comparison}
\end{table}
The \unitVar method assumes a fixed unit variance and predicts only the mean~\cite{sekar2020planning}, without distinguishing between aleatoric and epistemic uncertainty. \textit{EnsembleRSSM}~\cite{rafailov2021offline, seyde2022learning} trains an ensemble of transition models \(\dynamics_\theta\), using random sampling of ensemble indices at each step and estimating uncertainty via the variance of mean predictions. The \maxVar approach, defined as \(\max_k \left\| \Sigma_{\ensembleparam_k}(z_t, a_t) \right\|_F\), uses the maximum variance across the ensemble as a proxy for max aleatoric uncertainty~\cite{yu2020mopo, sun2023model}. Additionally, we evaluate a density-estimation method based on neural spline flows~\cite{durkan2019neural} trained on the learned latent space, which estimates the likelihood of either latent-action pairs \((\latent, \action)\) or latent states alone (\(\latent\)).

Overall, the JRD formulation achieves the best performance in both the balanced accuracy of the safety value function and the closed-loop evaluation. While the gap is less pronounced in the Dubins car setting, which is relatively simple and contains limited aleatoric uncertainty, methods that fail to distinguish between aleatoric and epistemic uncertainty consistently underperform. These approaches struggle to isolate epistemic uncertainty, which is critical for detecting OOD transitions that stem from limited training coverage. Density-based methods show higher false positive rates, showing limited effectiveness in modeling likelihood in high-dimensional latent spaces. In particular, latent-only density models exhibit the worst performance, frequently misclassifying in-distribution safe states as OOD. This is likely due to the latent dynamics model hallucinating overconfident predictions on OOD actions during imagination, highlighting the importance of transition-based OOD detection for the uncertainty-aware reachability analysis in imagination.

\subsection{Block Plucking}\label{sec:appendix_experiments_details_isaac}

\begin{table}[h]
\centering
\footnotesize
\renewcommand{\arraystretch}{1.2}
\begin{tabular}{rrr}
\toprule
\multicolumn{1}{c}{\textbf{Reward Term}} & 
\multicolumn{1}{c}{\textbf{Condition}} & 
\multicolumn{1}{c}{\textbf{Weight}} \\
\midrule
\texttt{success} & \texttt{stacked\_3on1} $\land$ $\lnot$\texttt{stacked\_2on1} $\land$ $\lnot$\texttt{stacked\_3on2} $\land$ $\lnot$\texttt{c2\_drop} & +10 \\
\texttt{failure} & \texttt{c2\_drop} $\lor$ \texttt{c3\_drop} & -10 \\
\texttt{check\_3on1} & Block $c_3$ is stacked on $c_1$ & +1 \\
\texttt{not\_stacked\_2on1} & Block $c_2$ is not stacked on $c_1$ & +1 \\
\texttt{not\_stacked\_3on2} & Block $c_3$ is not stacked on $c_2$ & +2 \\
\texttt{dist\_12} & Relative distance between $c_1$ and $c_2$ & +5 \\
\bottomrule
\end{tabular}
\vspace{0.05in}
\caption{Reward terms used for training the task policy. The goal is to extract the middle block ($c_2$) and place it on the base block ($c_1$) without collapsing the tower.}
\label{tab:reward_design}
\vspace{-0.2in}
\end{table} 

\paragraph{Task Policy Training.} The task policy is trained using DreamerV3~\cite{hafner2023dreamerv3} with dense reward signals. The environment consists of three blocks: \( c_1 \) (base block), \( c_2 \) (middle target block), and \( c_3 \) (top block). The goal is to extract \( c_2 \) from the tower and place it on top of \( c_1 \) without causing the tower to collapse. Table~\ref{tab:reward_design} shows the reward design for training the task policy. The action space is normalized to [-1, 1], and the task policy is modeled as a Gaussian distribution. During execution, only the mean is used, with a small additive noise sampled from \texttt{Uniform} [-0.02, 0.02]. Note that for training the world model and latent safety filter, only binary \texttt{failure} labels are used.

\paragraph{Experimental Setup} In Table~\ref{tab:isaac_result}, \textit{No Filter} refers to the base policy executed without any safety intervention. \baselineLatent uses only the learned failure margin function without explicitly modeling OOD failures. \safeOnly represents a setting where both the latent dynamics and safety filter are trained exclusively on successful demonstrations with 1500 trajectories, implicitly treating all failures as OOD. This setup aligns with prior approaches in failure detection and safety analysis that define the safe set based solely on successful demonstrations~\cite{castaneda2023distribution, xu2025can}.

\subsection{Jenga Experiments}\label{sec:appendix_experiments_details_jenga}

\begin{figure}[h]
\centering
    \includegraphics[width=0.5\linewidth]{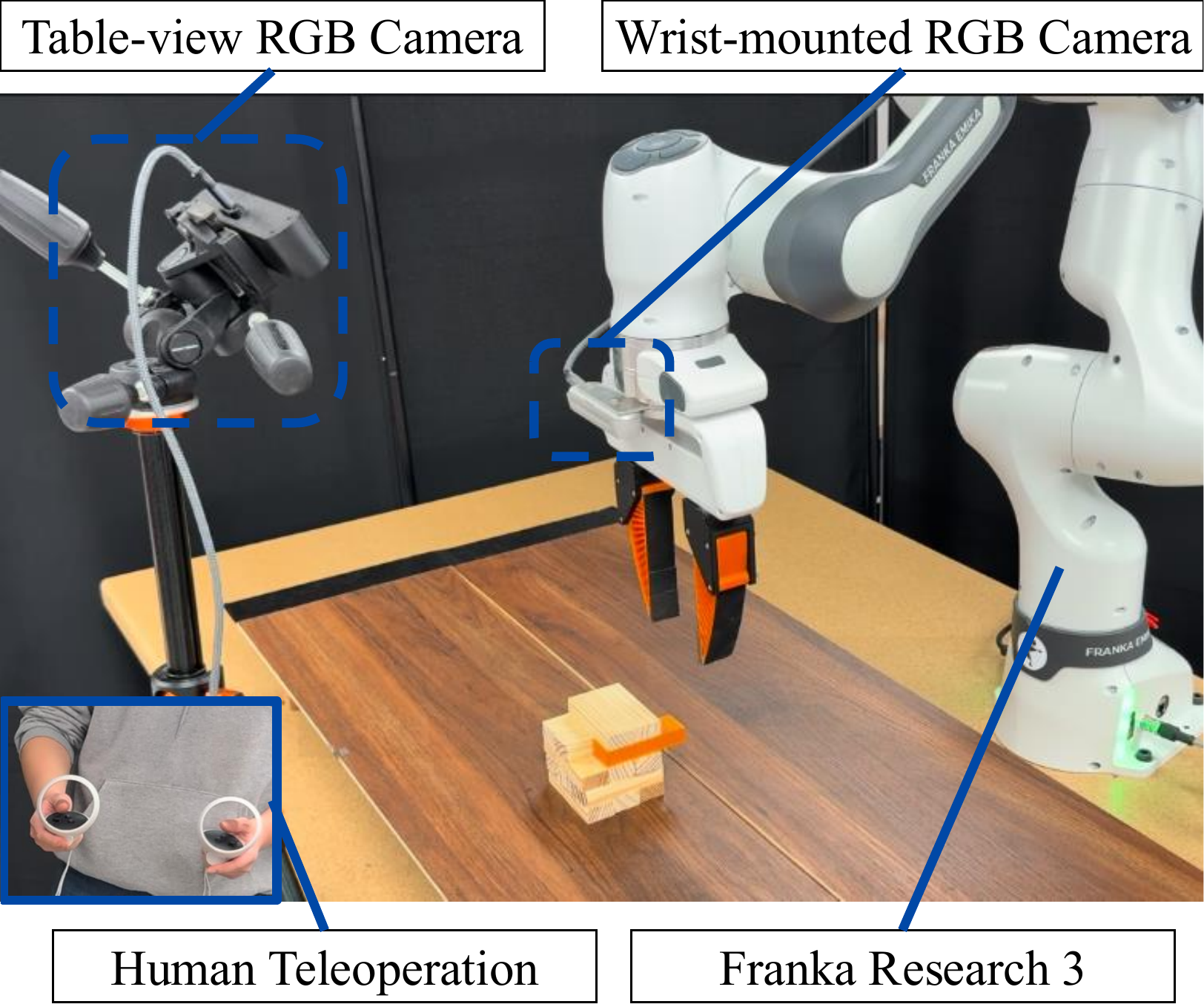}
    \caption{Setup for the Jenga experiments.}
    \label{fig:jenga_setup}
\end{figure}
\paragraph{Hardware Setup} Fig.~\ref{fig:jenga_setup} shows the setup. The fixed-base Franka Research 3 manipulator is equipped with a 3D-printed gripper~\cite{chi2024universal}. Two RGB cameras (third-person and wrist-mounted) capture 256×256 images at 15 Hz. It also takes 7D proprioceptive inputs (6D end-effector pose and gripper state) as input. A teleoperator uses a Meta Quest Pro to control the end-effector pose and gripper state. The robot must extract a target orange block from a tower and place it on top without causing collapse—a task characterized by high uncertainty due to complex contact dynamics and limited coverage in the offline dataset.

We keep the block configurations fixed, although minor variations in block positions naturally occur across trajectories. While the model cannot reliably filter block towers with entirely different configurations, it can still detect such settings as out-of-distribution (see Sec.~\ref{sec:failure-detection}). We believe that collecting more diverse demonstrations with varying configurations would enable the world model to build confidence across different setups and generalize more effectively to broader scenarios.

\begin{table}[h]
    \centering
    \small
    \renewcommand{\arraystretch}{1.2}{
        \begin{tabular}{cccc}
            \toprule
            \textbf{Method} & \textbf{Failure} $\downarrow$ & \textbf{Filtered}(\%) & \textbf{Model Loss} $\downarrow$\\
            \midrule
            \ours & 0.08 $\pm$ 0.08 &  0.19 & 28.49 $\pm$ 7.72\\
            \baselineLatent &  0.82 $\pm$ 0.11 & 0.08 & 33.25 $\pm$ 13.22\\
            \bottomrule
        \end{tabular}   
    }\vspace{0.1in} \caption{Open Loop Evaluation.}\label{tab:jenga_openloop} \vspace{-0.2in} 
\end{table}

\paragraph{Open-loop Rollout Experiments} For open-loop experiments, \ours uses a filtering threshold of \(\delta = 0.05\). In contrast, \baselineLatent exhibits inflated safety values due to overestimation of OOD actions. To enable effective filtering and ensure a fair comparison, we increase the threshold to \(\delta = 0.2\) for \baselineLatent, allowing it to identify the safe set more reasonably. However, even with the higher threshold, it still selects uncertain actions, as the elevated value estimates are attributed to the overestimation of these OOD actions. The detailed results are summarized in Table~\ref{tab:jenga_openloop}. The detailed results are summarized in Table~\ref{tab:jenga_openloop}. \baselineLatent still fails to intervene at the appropriate moment and exhibits higher model loss.

\section{Additional Results.}\label{sec:additional_results}

\subsection{How do dataset size and failure classifier performance affect the safety filter?}\label{sec:dubins_ablations}

\begin{table}[h]
    \centering
    \scriptsize
    \resizebox{0.7\textwidth}{!}{
    \renewcommand{\arraystretch}{1.05}
    \begin{tabular}{lcccccc}
    \toprule
    & \multicolumn{6}{c}{\textbf{Number of Random Trajectories}} \\
    \cmidrule{2-7}
    & \textbf{0} & \textbf{10} & \textbf{50} & \textbf{100} & \textbf{500} & \textbf{1000} \\
    \midrule
    Safe (\%)                     & 100.0 & 99.4 & 98.5 & 97.5 & 91.1 & 85.0 \\
    Unsafe (\%)                   & 0.0   & 0.6  & 1.5  & 2.5  & 8.9  & 15.0 \\
    Failure (\%)                  & 0.0   & 0.4  & 1.1  & 1.9  & 6.7  & 11.3 \\
    Failure Classifier (Acc.)     & 0.82  & 0.90 & 0.91 & 0.89 & 0.96 & 0.97 \\
    \midrule
    \textbf{B.Acc (\baselineLatent)}        & 0.50 & 0.79 & 0.84 & 0.75 & 0.97 & 0.97 \\
    \textbf{B.Acc (\ours)}         & 0.84 & 0.93 & 0.93 & 0.92 & 0.97 & 0.97 \\
    \bottomrule
    \end{tabular}}
    \vspace{0.05in}
    \caption{Dataset and failure classifier configurations for ablations on the Dubins Car.}
    \label{tab:dubins_expert_ablation_result}
    \vspace{-0.2in}
\end{table}

Our uncertainty-aware safety filter relies on two types of failure sets: (\textit{i}) the known failure set, derived from labeled failure data, and (\textit{ii}) the OOD failure set, which accounts for distributional shift and epistemic uncertainty. As the dataset size increases, the latent world model and failure classifier become more accurate, reducing the size and impact of the OOD failure set. In contrast, with smaller datasets, large regions of the state space remain uncovered, making the OOD failure set crucial for robust safety filtering.

To investigate how dataset size impacts the learned safety filter, we perform an ablation study varying the number of random trajectories. Following the Dubins Car setup in Sec.~\ref{sec:dubins}, we construct a dataset consisting of 1000 expert trajectories that never enter the ground-truth unsafe set, along with a varying number of random trajectories, some of which do enter unsafe or failure regions. Using privileged state information, we train a failure margin function to approximate the known failure set. The number of random trajectories is varied across $\{0, 10, 50, 100, 500, 1000\}$. When fewer random trajectories are included, the dataset is biased toward safe states, increasing epistemic uncertainty near the unsafe set and reducing the accuracy of the failure classifier, thereby necessitating the inclusion of OOD failure modeling to capture risk in unexplored regions.

Table~\ref{tab:dubins_expert_ablation_result} summarizes the dataset statistics and failure classifier performance across different dataset sizes. As the number of random trajectories increases, a larger portion of the failure region is covered, improving the failure classifier's accuracy. 

\begin{figure}[h]
    \centering
    \includegraphics[width=0.9\linewidth]{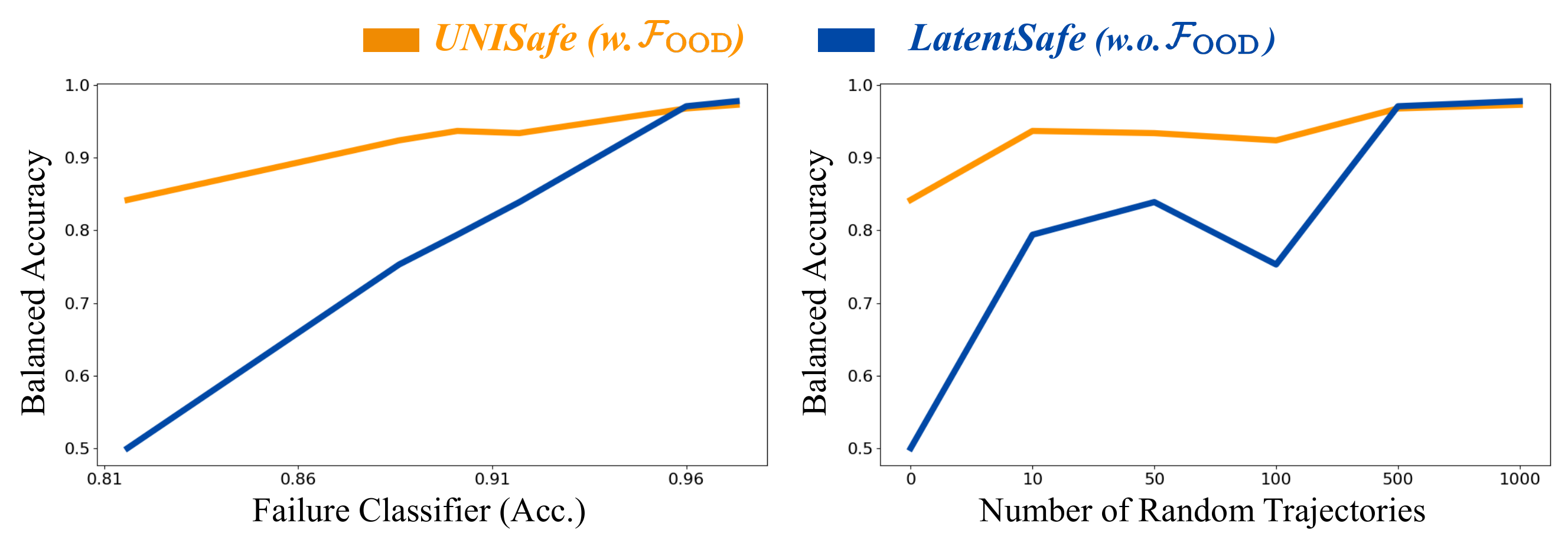}
    \caption{\ours can synthesize a much more robust safety filter even under an unreliable world model. When trained on smaller datasets with an ineffective failure classifier, \baselineLatent results in an unreliable safety filter, whereas \ours remains robust by explicitly incorporating OOD failures.}
    \label{fig:dubins_ablation}
\end{figure}

We then evaluate the performance of the safety value function trained under each dataset configuration. The results indicate that incorporating OOD failures via uncertainty quantification is particularly beneficial when the dataset does not adequately cover the true failure set. When fewer random trajectories are available, safety filters trained without OOD modeling exhibit significantly degraded performance due to unmodeled epistemic uncertainty in the dynamics model. Figure~\ref{fig:dubins_ablation} visualizes the accuracy of the learned safety value function with respect to the ground-truth failure set. The results demonstrate that integrating OOD detection yields substantially more robust performance, especially when the failure classifier is weak and the dataset is small. In contrast, the baseline approach, trained without OOD detection, fails to maintain safety in such challenging conditions.

\subsection{The safety filter reliably safeguards diverse base policies.}

\begin{table}[h]
    \centering
    \scriptsize
    \setlength{\tabcolsep}{4pt}
    \renewcommand{\arraystretch}{1.0}
    \begin{tabular}{ccccccccc}
        \toprule
         & $\policyTask$ & \textbf{Method} & \textbf{Safe Success ($\uparrow$)} & \textbf{Failure ($\downarrow$)} & \textbf{Incompletion} & \textbf{Filtered} ($\%$) & \textbf{Seq. Length} & \textbf{Model Error ($\downarrow$)}\\ 
         \midrule
         \multirow{9}{*}{\rotatebox{90}{\textbf{Normal}}} 
        & \multirow{4}{*}{\rotatebox{90}{Dreamer}} & No Filter & 0.58 & 0.41 & 0.01 & 0.0 ± 0.0 & 61.2 ± 45.7 & 59.3 ± 3.3 \\ 
        & & \safeOnly & 0.71 & 0.28 & 0.01 & 13.5 ± 3.5 & 70.25 ± 54.5 & 46.9 ± 2.6 \\
        & & \baselineLatent & 0.68 & 0.30 & 0.01 & 7.2 ± 2.6 & 70.01 ± 55.8 & 60.2 ± 4.7 \\ 
        & & \ours & 0.72 & 0.20 & 0.08 & 37.7 ± 6.7 & 95.7 ± 90.2 & 43.1 ± 1.2 \\ 
        \cmidrule(lr){2-9}
        & \multirow{4}{*}{\rotatebox{90}{\shortstack{Diffusion\\Policy}}} & No Filter & 0.52 & 0.44 & 0.04 & 0.0 ± 0.0 & 87.3 ± 74.6 & 89.8 ± 20.9 \\ 
        & & \safeOnly & 0.50 & 0.38 & 0.12 & 26.6 ± 3.8 & 126.9 ± 101.0 & 85.1 ± 20.8 \\
        & & \baselineLatent & 0.42 & 0.51 & 0.07 & 9.3 ± 2.6 & 93.5 ± 82.1 & 70.7 ± 18.1 \\ 
        & & \ours & 0.57 & 0.15 & 0.28 & 37.6 ± 5.5 & 150.4 ± 120.0 & 48.3 ± 5.5 \\ 
        \midrule
        \multirow{9}{*}{\rotatebox{90}{\textbf{Hard}}} 
        & \multirow{4}{*}{\rotatebox{90}{Dreamer}} & No Filter & 0.48 & 0.52 & 0.00 & 0.0 ± 0.0 & 53.03 ± 43.04 & 44.5 ± 4.8 \\ 
        & & \safeOnly & 0.47 & 0.46 & 0.07 & 50.7 ± 8.5 & 68.9 ± 77.3 & 45.5 ± 9.2 \\ 
        & & \baselineLatent & 0.51 & 0.49 & 0.00 & 15.9 ± 3.9 & 60.3 ± 52.7 & 44.9 ± 10.9 \\ 
        & & \ours & 0.64 & 0.22 & 0.14 & 40.0 ± 6.1 & 103.5 ± 95.0 & 37.3 ± 2.5 \\ 
        \cmidrule(lr){2-9}
        & \multirow{4}{*}{\rotatebox{90}{\shortstack{Diffusion\\Policy}}} & No Filter & 0.17 & 0.58 & 0.25 & 0.0 ± 0.0 & 179.4 ± 91.5 & 79.6 ± 20.0 \\ 
        & & \safeOnly & 0.29 & 0.53 & 0.18 & 22.9 ± 2.5 & 163.9 ± 88.0 & 82.7 ± 19.7 \\
        & & \baselineLatent & 0.18 & 0.62 & 0.20 & 6.4 ± 0.9 & 183.5 ± 88.7 & 75.2 ± 22.5 \\ 
        & & \ours & 0.38 & 0.31 & 0.31 & 21.9 ± 2.0 & 179.8 ± 91.3 & 43.9 ± 13.5 \\ 
        \bottomrule
    \end{tabular}
    \vspace{0.05in}
    \caption{Additional result on block plucking in simulation environments.}
    \label{tab:isaac_result_addtional}
    \vspace{-0.2in}
\end{table}

\begin{figure}[h]
    \centering
    \includegraphics[width=1.0\linewidth]{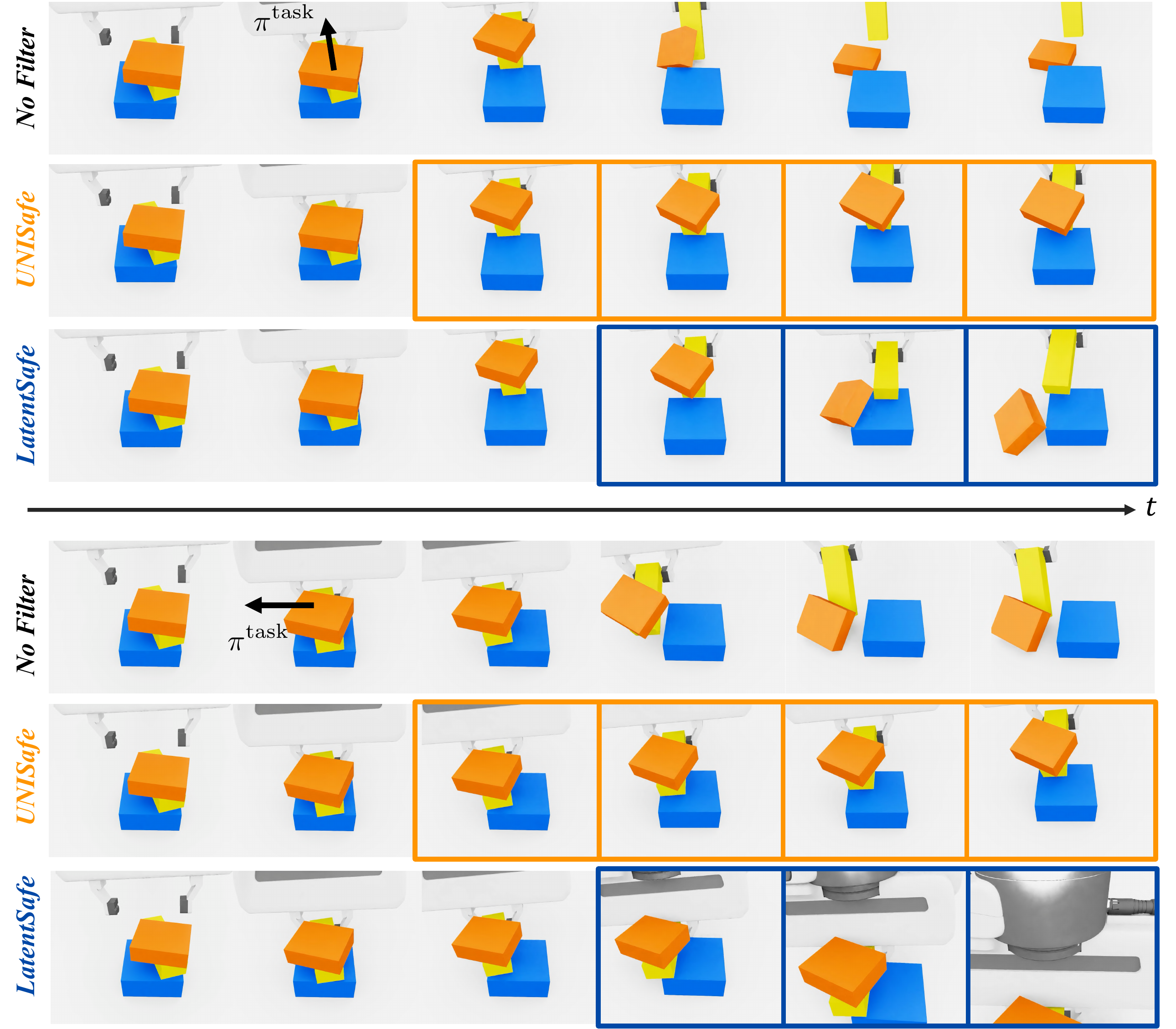}
    \caption{Qualitative results on the \textit{Hard} setting. The task policy attempts to pick the block but naively pulls it in one direction, causing failures where the orange block falls. \ours accurately detects the boundary of the unsafe set and prevents failures caused by blocks falling with momentum. The safety policy reliably corrects the task policy’s actions by proposing safe, in-distribution alternatives, keeping the block stable. In contrast, \baselineLatent detects the unsafe set too late and eventually proposes abrupt, unsafe actions that lead to failure.}
    \label{fig:isaac_hard}
    \vspace{-0.1in}
\end{figure}

For a more thorough evaluation, we additionally consider a \textit{Hard} setting (see Fig.~\ref{fig:isaac_hard}), which varies the block size, weight, and friction to allow for a more comprehensive assessment. As nominal task policies, we evaluate (1) DreamerV3\cite{hafner2023dreamerv3}, trained online with a dense reward signal, and (2) Diffusion Policy~\cite{chi2023diffusion}, an imitation learning trained on 200 safe trajectories. Table~\ref{tab:isaac_result_addtional} shows that \ours consistently minimizes failure rates and model errors compared to the baselines.

\subsection{Is model-based imagination or explicit uncertainty quantification essential?}\label{sec:appendix_conservative} We perform ablations to evaluate the necessity of the two core components in our uncertainty-aware latent-space reachability framework: (1) a \textit{latent dynamics model} for reachability analysis in imagination and (2) explicit \textit{epistemic uncertainty quantification} for preventing distributional shift.

\paragraph{Conservative Q-Learning} In the model-free offline RL setting,  policies can be learned solely from offline datasets, without learning a learned dynamics model. To address value overestimation on out-of-distribution actions, Conservative Q-Learning (CQL)~\cite{kumar2020conservative} introduces a conservative training objective that penalizes high Q-values for unseen or randomly sampled actions. This is achieved by augmenting the standard Bellman error with a behavioral-cloning-style regularization term, which constrains the Q-values for out-of-distribution actions while preserving those associated with in-distribution actions:
\begin{equation}\label{eq:bc-loss}
\min_Q \, \alpha \, \mathbb{E}_{\latent \sim \mathcal{D}} \left[ 
\log \sum_{\action} \exp(Q(\latent, \action)) 
- \mathbb{E}_{\action \sim \hat{\pi}_\beta(\action \mid \latent)} \left[ Q(\latent, \action) \right] 
\right],
\end{equation}
where \(\hat{\pi}_\beta\) denotes the behavior policy from the offline dataset. We apply this conservative loss to train the safety filter. The safety value function and policy are learned on top of the latent representation space using offline transitions without relying on model-based imagination.

This conservatism principle can also be extended to the model-based setting without requiring explicit uncertainty estimation. COMBO~\cite{yu2021combo} combines offline transitions with model-generated rollouts to train a value function and regularizes Q-values on out-of-support state-action pairs generated by the model. In our case, we adopt the same conservative objective from Eq.~\ref{eq:bc-loss}, applying it to imagined latent transitions produced by the learned latent dynamics model, thus eliminating the need for explicit epistemic uncertainty quantification.

\paragraph{Analysis} One limitation of the conservative objective is that, by directly penalizing the value function, it introduces bias into the learned safety values. As a result, the value function can no longer reliably serve as a level-set representation for identifying the unsafe set via its zero sublevel set. To address this, we use a calibration dataset to select a value threshold $\delta$ that best separates safe and unsafe states in practice.

Quantitative results of this ablation study are provided in Table~\ref{tab:isaac_result}. While learning a safety monitor and policy without model-based imagination is feasible, its effectiveness is limited in high-dimensional visual manipulation settings where the offline dataset may not adequately cover the state space, restricting the quality of reachability approximation. Furthermore, not using uncertainty quantification and relying solely on the conservative loss results in overly conservative filtering behavior. These results suggest that although conservative objectives are effective in standard offline RL tasks that aim to maximize expected returns, they are less suitable for safety analysis. Safety-critical applications require accurate specification of the safe and unsafe sets, which conservative regularization alone fails to guarantee.

\subsection{Can uncertainty-penalized offline RL ensure the safety of the task policy?}  
Offline reinforcement learning (RL) methods frequently incorporate uncertainty estimation to enhance safety and robustness when learning task policies from static datasets. To assess the effectiveness of uncertainty-penalized policy learning in this context, we train a task policy using an offline model-based reinforcement learning (MBRL) framework (\texttt{LOMPO}~\cite{rafailov2021offline}) that employs a latent dynamics model for image-based control. It formulates an uncertainty-penalized POMDP, where an ensemble of RSSMs is used to estimate epistemic uncertainty and penalize transitions softly accordingly during policy optimization. We investigate this with the visual manipulation tasks in simulation.

In contrast to the safety filter, which relies learned failure margin function, we learn dense task-relevant rewards \(\bar{r}_{\theta}(\latent_t, \action_t)\) for the task policy training. To ensure the penalty term accurately reflects epistemic uncertainty, we employ JRD-based uncertainty quantified by ensembles (Sec.~\ref{sec:uq-wm}).  The resulting reward function used for training is defined as: \(r_t(\latent_t, \action_t) = \bar{r}_{\theta}(\latent_t, \action_t) - \lambda \disagreement(\latent_t, \action_t),\) where \(\lambda = 0.5\) controls the strength of the uncertainty penalty.

\begin{table}[h]
    \centering
    \scriptsize
    \setlength{\tabcolsep}{4pt}
    \renewcommand{\arraystretch}{1.2}
    \begin{tabular}{ccccccccc}
        \toprule
         & $\policyTask$ & \textbf{Safety Filter} & \textbf{Safe Success} & \textbf{Failure} & \textbf{Incompletion} & \textbf{Filtered} ($\%$) & \textbf{Seq. Length} & \textbf{Model Error}\\ 
         \toprule
        \multirow{2}{*}{Normal} & \multirow{2}{*}{\texttt{LOMPO}~\cite{rafailov2021offline}} & No Filter & 0.36 & 0.63 & 0.01 & 0.0 ± 0.0 & 66.3 ± 54.2 & 79.1 ± 4.1 \\ 
        & & \ours & 0.41 & 0.26 & 0.33 & 60.3 ± 6.8 & 154.1 ± 122.9 & 54.3 ± 8.8 \\ 
        \cmidrule(lr){1-9}
        \multirow{2}{*}{Hard} & \multirow{2}{*}{\texttt{LOMPO}~\cite{rafailov2021offline}} & No Filter & 0.05 & 0.95 & 0.00 & 0.0 ± 0.0 & 83.7 ± 50.9 & 71.2 ± 16.3 \\ 
        & & \ours & 0.41 & 0.36 & 0.23 & 41.7 ± 4.9 & 143.5 ± 106.5 & 32.7 ± 10.3 \\
        \bottomrule
     \end{tabular}
     \vspace{0.05in}
    \caption{Rollout results with offline learned task policy following \texttt{LOMPO} with the pessimistic MDP.}
    \label{tab:lompo_result}
    \vspace{-0.1in}
\end{table}
Table~\ref{tab:lompo_result} presents the experimental results. The uncertainty-penalized policy trained via LOMPO exhibits high failure rates and accumulates large model errors during rollouts. It exhibits limited performance, especially in the Hard setting, suggesting that a soft uncertainty penalty in offline MBRL is insufficient for safety. Using the same offline dataset, we instead train a safety filter that explicitly incorporates uncertainty and use it to filter the task policy learned by \texttt{LOMPO}. While this approach does not guarantee zero failure, it significantly reduces the failure rate and also lowers the model error. These results suggest that an uncertainty-aware safety filter learned from offline data is more effective at ensuring safety than directly penalizing uncertainty during task policy optimization. Additionally, when uncertainty penalties are omitted entirely during offline policy learning, the resulting policy fails to perform meaningful behavior, primarily due to value overestimation, a well-documented challenge in offline RL.

\end{document}